\pdfoutput=1
\documentclass[10pt,twocolumn,letterpaper]{article}

\usepackage[pagenumbers]{cvpr}

\usepackage{cuted}
\usepackage{currfile}
\usepackage{caption}
\usepackage{placeins}
\usepackage{stfloats}
\usepackage{algorithm}
\usepackage{algpseudocode}

\makeatletter
\AtBeginDocument{%
  \@ifpackageloaded{lineno}{%
    \IfHookExistsTF{build/column/after}{%
      \RemoveFromHook{build/column/before}[lineno]%
      \AddToHook{build/column/after}[heatgen/lineno-column]{%
        \setbox\@outputbox\vbox{%
          \boxmaxdepth\@maxdepth
          \protected@write\@auxout{}{%
            \string\def\string\@LN@column{\if@firstcolumn1\else2\fi}}%
          \box\@outputbox}}%
    }{}%
  }{}%
}
\makeatother

\usepackage{pgfplots}
\pgfplotsset{compat=1.18}
\usetikzlibrary{arrows.meta}
\usepackage{pifont}
\newcommand{\cmark}{\ding{51}}
\newcommand{\xmark}{\ding{55}}
\usepackage{subcaption}
\newcommand{\twoline}[2]{\begin{tabular}[b]{@{}c@{}}#1\\#2\end{tabular}}

\definecolor{cvprblue}{rgb}{0.21,0.49,0.74}
\usepackage[pagebackref,breaklinks,colorlinks,allcolors=cvprblue]{hyperref}

\title{HeatGen: Intrinsic Texture Diffusion with Heat-Closed Chart Tokens}
\author{Zhewen He \qquad Junyi Hu \qquad Yi Fang\thanks{Corresponding author.}\\[4pt]
New York University Abu Dhabi
}

\begin{document}
\maketitle

\begin{abstract}
Meshes from image-to-3D generators and scanners rarely have a usable UV atlas, and they are full of thin shells and touching parts. Multi-view baking must inpaint what the views miss, and native 3D latents use Euclidean neighbourhoods that blend the two sides of a thin shell. We present HeatGen, a texture diffusion model whose state lives on the surface. Its \emph{chart tokens} are the inner products of the texture with Gaussian-windowed Fourier--Bessel atoms on overlapping geodesic charts. They are local and exactly steerable within each chart, and on flat charts the atoms are closed under heat flow. The tokens of a heat-blurred texture are therefore inner products with analytically blurred atoms. This yields a coarse-to-fine forward process that is consistent with surface heat flow and needs no eigenbasis. A gauge-equivariant transformer denoises the tokens. On image-to-mesh texturing, HeatGen matches or improves on strong baselines on overall metrics and is clearly better in occluded and thin regions, where it roughly halves the colour leaking between separable layers. The gains persist against other surface states trained with the same data and conditioning, and the forward process matches mesh-resolution heat diffusion at a fraction of its sampling cost.
\end{abstract}

\section{Introduction}
\label{sec:intro}

\begin{figure}[t]
\centering
\includegraphics[width=\linewidth]{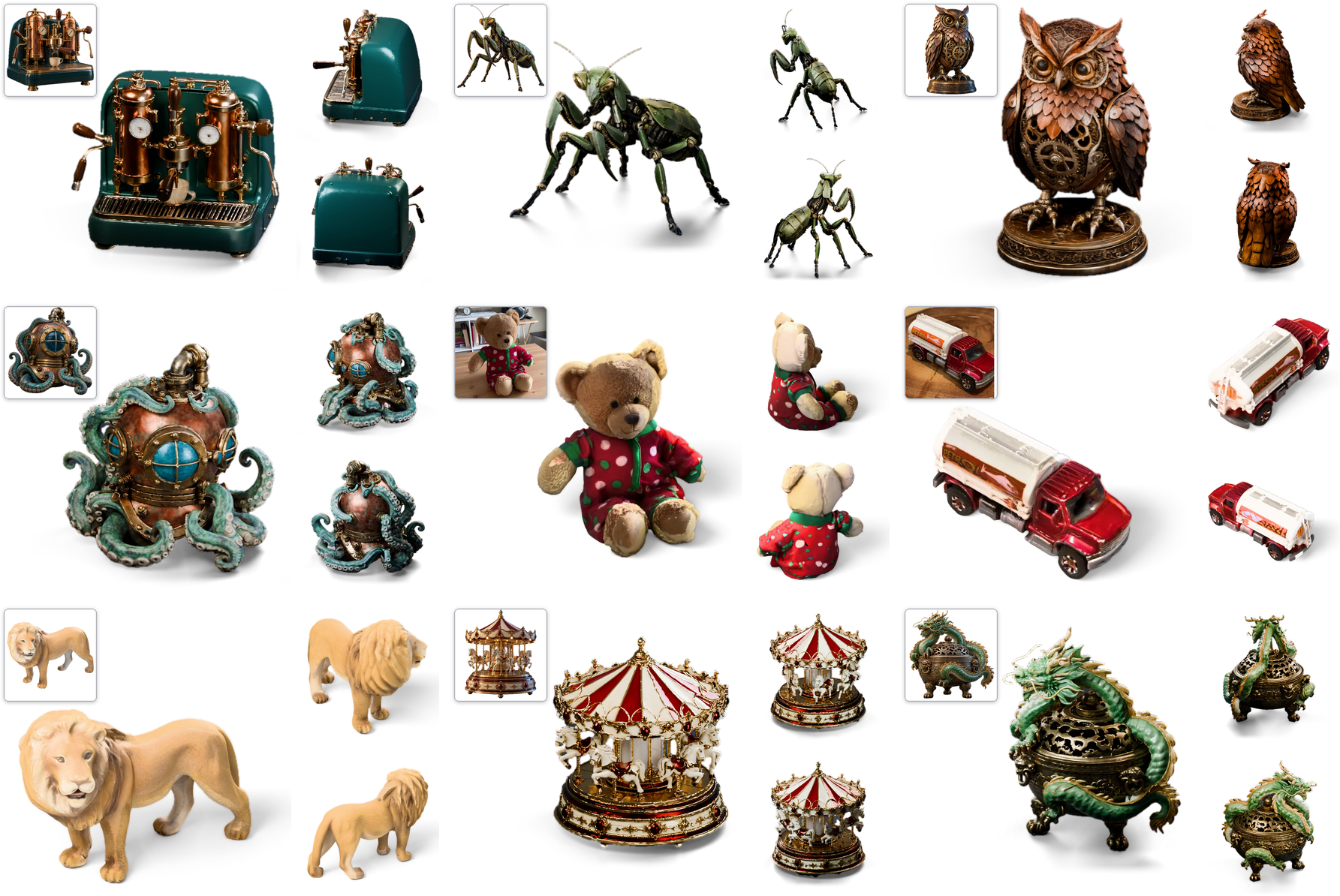}
\caption{HeatGen textures arbitrary meshes from a single image (inset) with steerable, heat-closed chart tokens. Large: render near the input view; small: side and unseen back. None of these objects appear in Fig.~\ref{fig:qual}.}
\label{fig:teaser}
\end{figure}

A growing share of 3D assets comes from image-to-3D generators~\cite{trellis,trellis2,hunyuan} and scanning. Such meshes have no usable UV atlas, are often non-manifold, and contain thin structures: cloth, leaves, closed shells from SDF extraction, and touching parts. \emph{View-based} texturing~\cite{texture,text2tex,paint3d,mvadapter,hunyuan} inherits strong image priors, but must inpaint whatever the views miss and leaves seams. \emph{UV-space} generators~\cite{pointuv,texgen} need an atlas. \emph{Native 3D} latents~\cite{trellis2,natex,lafite,textrix,unitex} are seam-free but use Euclidean neighbourhoods, so two sheets closer than the latent resolution blend (Fig.~\ref{fig:qual}). \emph{Intrinsic} models~\cite{uv3ted,double,field,mdf} work on the surface, but their state lives on vertices or points, so resolution follows the sampling, and spectral variants need Laplacian eigenvectors.

We ask what a local, mesh-independent surface state for diffusion should be. Following frequency-domain image diffusion~\cite{dctdiff,bdm}, we cover the surface with overlapping geodesic charts and record, on each chart, the inner products of the texture with \emph{Gaussian--Fourier--Bessel atoms}: a Gaussian window times circular harmonics, where the ring sets the frequency and the order $m$ the angular structure. The coefficients of one chart form a \emph{chart token}. Two properties make chart tokens a good diffusion state. They are \emph{steerable}: rotating a chart's frame by $\phi$ multiplies each order-$m$ coefficient by $e^{\mathrm im\phi}$, so a network can be equivariant to every chart's frame without a global direction field~\cite{hsn,gem}. They are \emph{heat-closed}: under heat flow an atom stays an atom of the same order with a wider window and lower frequency, the imaginary-time analogue of Bessel--Gauss beams~\cite{bessel}. Since heat flow is self-adjoint, blurring the texture along the surface amounts to taking inner products with blurred atoms, chart by chart. Within the tokens' band and up to curvature error, this gives the mean of blurring diffusion~\cite{bdm,ihdm} in the Laplacian eigenbasis, without ever forming it.

\emph{HeatGen} builds on these tokens a cross-shape, image-conditioned texture generator with a gauge-equivariant transformer. Our contributions are:
\begin{itemize}
\item \textbf{Chart tokens}, a local, exactly steerable surface state with a well-conditioned decoder, whose resolution is set by anchors rather than the triangulation, and a gauge-equivariant denoiser for it. Under matched data, budget and conditioning, they improve on the strongest alternative surface state, most of all where the mesh changes.
\item \textbf{An efficient heat-consistent forward process}: closed-form blurred atoms at fine scales and anchor-graph heat at coarse ones, with no eigenbasis or mesh-resolution state. It beats tuned per-ring schedules and matches mesh-resolution heat at a fraction of the sampling cost (Sec.~\ref{sec:exp:forward}).
\item \textbf{Image-to-mesh results} that match or improve on the strongest systems overall, with clear gains in occluded and thin regions, which are common in generated meshes (Sec.~\ref{sec:exp:i2m}, \ref{sec:exp:thin}).
\end{itemize}

\section{Related Work}
\label{sec:related}

\paragraph{Texture generation with 2D priors.}
Most texture generators lift pretrained image diffusion models to 3D. Methods paint and project one view at a time~\cite{texture,text2tex}, or generate consistent multi-view images and bake them~\cite{paint3d,mvadapter,hunyuan,seqtex,im2surftex}. GOATex~\cite{goatex} textures occluded layers separately by ray-cast visibility. Hitem3D~2.0~\cite{hitem3d} combines multi-view guidance with a native texture model. They inherit strong image priors, but their output is assembled from projections, so occluded regions must be completed. We generate on the surface and use the image only as conditioning.

\paragraph{Native texture generation in UV and 3D space.}
Point-UV Diffusion~\cite{pointuv} and TEXGen~\cite{texgen} diffuse in UV space and therefore require an atlas. DirectUV~\cite{directuv} and texture-space material diffusion~\cite{tsmd} also operate on a given parameterization. Volumetric and point-based models avoid the atlas. TRELLIS~\cite{trellis,trellis2} generates structured sparse-voxel latents, NaTex~\cite{natex} diffuses latent colour point clouds, and LaFiTe~\cite{lafite}, TEXTRIX~\cite{textrix}, UniTEX~\cite{unitex}, TexSpot~\cite{texlet} and the concurrent TaoTex~\cite{taotex} use latent fields, attribute grids, texture functions and surface point latents. TexSpot encodes local texture patches into Texlet latents with a learned decoder; chart tokens are analysis coefficients with a fixed linear decoder and exact steerability. They are seam-free, but their neighbourhoods are Euclidean, so surfaces closer than the latent resolution share appearance.

\paragraph{Generative models on surfaces.}
Manifold Diffusion Fields~\cite{mdf} generate functions on a manifold using Laplace--Beltrami eigenfunctions as coordinates. UV3-TeD~\cite{uv3ted} generates coloured point clouds across shapes and propagates information by spectral heat diffusion with 128 eigenvectors. DoubleDiffusion~\cite{double} denoises per-vertex colours with DiffusionNet-style heat blocks~\cite{diffusionnet}, and Mat\'ern noise~\cite{matern} makes flow matching on meshes agnostic to the triangulation. There, heat is a propagation operator or a noise prior, and the state lives on vertices or points. Field Latent Diffusion~\cite{field} is closest in spirit. It encodes texture as tangent-vector latents at vertices and denoises them with isometry-equivariant field convolutions, but it is trained on a single mesh. We share its tangent frames and equivariance, but our state is an explicit analysis whose size is set by anchors, our model is trained across shapes with image conditioning, and our forward process follows from heat closure.

\paragraph{Frequency-domain and heat-based diffusion.}
DCTdiff~\cite{dctdiff} diffuses image DCT coefficients, and wavelet score models~\cite{wavelet} factor generation across scales. Inverse heat dissipation~\cite{ihdm}, blurring diffusion~\cite{bdm}, blur--noise mixtures~\cite{warm} and heat-dissipation flow matching~\cite{hdfm} corrupt images by heat flow. Blurring diffusion is Gaussian diffusion with a frequency-dependent schedule that is diagonal in the Laplacian eigenbasis, and spectral noise schedules design that dependence explicitly~\cite{spectral}. Heat flow on meshes can be applied without an eigendecomposition through Chebyshev or Krylov methods~\cite{cheb,expmv}. We make heat-based corruption local and eigenbasis-free on surfaces.

\paragraph{Local charts, steerable atoms and gauge equivariance.}
Lapped textures~\cite{lapped} cover a surface with overlapping patches, and Gabor noise~\cite{gabor} offers local spectral control. Circular-harmonic and Fourier--Bessel bases make planar features and filters steerable~\cite{fbpca,rotdcf,harmonic,steerable}, finite Fourier--Bessel transforms can be kept closed under convolution on disks~\cite{fbclosure}, and concurrent notes construct Gaussian-windowed Fourier--Bessel wavelets~\cite{fbwavelets}. Harmonic Surface Networks~\cite{hsn} carry circular-harmonic features on tangent planes and align them by parallel transport. TextureNet~\cite{texturenet} convolves surface textures in local parametrizations oriented by a 4-RoSy field, which fixes orientation only up to quarter turns. Gauge-equivariant mesh networks follow the same principle with convolutions and attention~\cite{gem,get,eman}. Heat Kernel Textures~\cite{hktex} represent appearance with fitted anisotropic heat kernels, and batched local unfolding makes such kernels fast without mesh eigensystems~\cite{localhk}. We reuse the unfolding and connection angles of~\cite{localhk}. Unlike these discriminative networks and fitted representations, we use the analysis coefficients of steerable atoms as the state of a generative model and exploit their closure under heat flow.

\section{Method}
\label{sec:method}

\begin{figure*}[t]
\centering
\includegraphics[width=\textwidth]{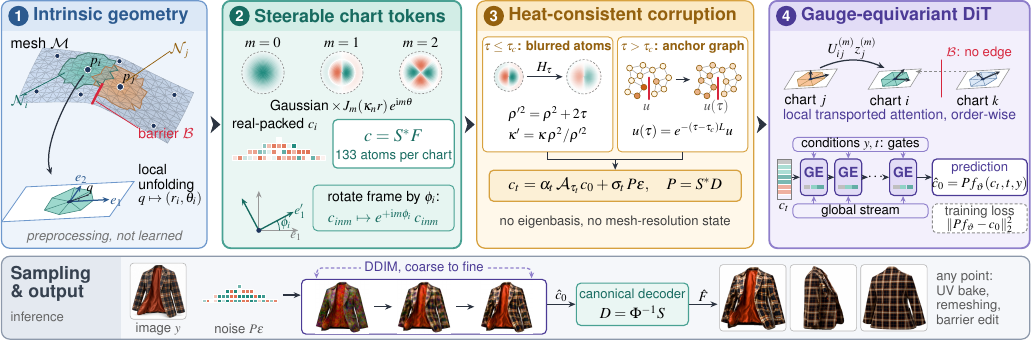}
\caption{Pipeline overview. (1) Anchors are unfolded into geodesic charts that never cross barriers $\mathcal B$. (2) Chart tokens are inner products with Gaussian--Fourier--Bessel atoms; a frame rotation multiplies order $m$ by a phase. (3) The forward process blurs by surface heat flow and adds noise: closed-form blurred atoms up to $\tau_c$, anchor-graph heat beyond. (4) A gauge-equivariant transformer predicts clean tokens; the decoder $D=\Phi^{-1}S$ evaluates the texture anywhere. Bottom: sampling for the jacket of Fig.~\ref{fig:qual}.}
\label{fig:pipeline}
\end{figure*}

Given a mesh $\mathcal M$ and a conditioning image $y$, HeatGen generates a colour field $F:\mathcal M\to\mathbb R^C$. We use $C=3$ for albedo. The mesh may be a triangle soup or non-manifold. In that case we work on its tufted cover~\cite{robustlap}, which is edge-manifold and shares its geometry. Fig.~\ref{fig:pipeline} summarizes the pipeline.

\subsection{Charts}
\label{sec:charts}

We place $N$ anchors $\{p_i\}$ with mean spacing $h$ by farthest-point sampling and Lloyd relaxation. Each is unfolded into the tangent plane by the batched hinge unfolding of~\cite{localhk} up to radius $r_{\rm c}=3.4h$, never crossing an optional barrier set $\mathcal B$ (\eg mask boundaries). A surface point $q$ in chart $i$ receives polar coordinates $(r_i(q),\theta_i(q))$. Composed hinge rotations give connection angles $\phi_{ij}$ between overlapping charts (a discrete parallel transport). Chart frames are arbitrary; the network is equivariant to them.

Because charts follow the surface, sheets that are close in space but far apart along it never share a chart. On raw generator meshes the barrier set is empty, and separation comes from connectivity alone. Barriers are added only for editing, on mask boundaries, or where a user marks a seam that should not be crossed.

\subsection{Chart Tokens}
\label{sec:tokens}

\paragraph{Atoms.} On chart $i$ we define Gaussian--Fourier--Bessel atoms
\begin{equation}
\psi_{i,n,m}(q)=Z_{n,m}^{-1}\,g_\rho\big(r_i(q)\big)\,J_m\big(\kappa_n r_i(q)\big)\,e^{\mathrm i m\theta_i(q)},
\label{eq:atom}
\end{equation}
where $g_\rho(r)=e^{-r^2/2\rho^2}$, $J_m$ is the Bessel function of order $m$, and $Z_{n,m}$ normalizes each atom to unit $L^2$ norm. Ring $n=0$ has $\kappa_0=0$ and carries the local mean. Rings $n=1,\dots,6$ have $\kappa_n=2n/\rho$ and orders $|m|\le 3n$. With $\rho=h/2$ this gives 133 atoms per chart, covering frequencies up to $k_{\max}\approx 25/h$.

\paragraph{Analysis state.} A chart token is the vector of analysis coefficients
\begin{equation}
c_{i,n,m}=\int_{\mathcal M}\overline{\psi_{i,n,m}(q)}\,F(q)\,\mathrm dA(q),\qquad c=S^*F,
\label{eq:analysis}
\end{equation}
which we store as real numbers per colour channel. No fitting is involved: $c$ is unique and linear in $F$, and $c_i$ depends only on $F$ within chart $i$. The state has $N\times133\times C$ entries for every mesh. It covers about $50$ texels per anchor and channel, for example about $900^2$ at $N=16{,}384$.

\paragraph{Steerability.} Rotating the frame of chart $i$ by $\phi$ maps $\theta_i\mapsto\theta_i-\phi$, so every atom gains a factor $e^{-\mathrm im\phi}$ and, by Eq.~\eqref{eq:analysis},
\begin{equation}
c_{i,n,m}\mapsto e^{\mathrm i m\phi}\,c_{i,n,m}.
\label{eq:gauge}
\end{equation}
This holds exactly, because the angular dependence is analytic and no angular sampling is involved.

\paragraph{Decoding.} Let $S$ be the synthesis operator $c\mapsto\sum c_{i,n,m}\psi_{i,n,m}$ and $\Phi=SS^*$ the frame operator. On the span $V$ of the atoms (Supp.~\ref{supp:frame}), $\Phi$ is bounded above and below with $B/A=2.86$ on a regular anchor grid, computed by Fourier fiberization, and has a median of 2.94 for our irregular anchors (Supp.~\ref{supp:checks}). The decoder
\begin{equation}
D=\Phi^{-1}S,\qquad P=S^*D,
\label{eq:decoder}
\end{equation}
recovers $F$ from $c$ by a few conjugate-gradient iterations. $P$ projects any coefficient vector onto $\mathrm{range}(S^*)$, the set of valid tokens. \paragraph{Redundancy and projection.} A basis that is both spatially local and exactly steerable cannot be orthogonal, since overlapping windows share content. Chart tokens are therefore redundant, with about $2.7$ atoms per resolved texel (a different quantity from the conditioning ratio $B/A$), and about 63\% of coefficient directions do not correspond to any texture. Noise or predictions in those directions would decode to nothing and waste denoiser capacity, so we project both noise and network outputs with $P$. Diffusion then runs on the valid subspace, where the decoder is exact, and DDIM remains valid (Sec.~\ref{sec:training}).

\begin{figure}[t]
\centering
\includegraphics[width=\linewidth]{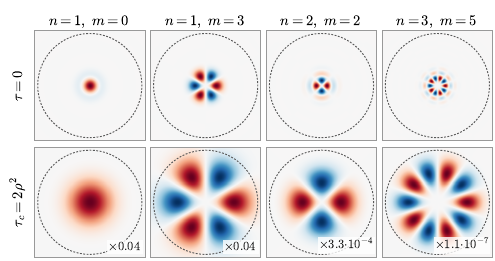}
\caption{Heat closure of Gaussian--Fourier--Bessel atoms (real parts, each panel normalized; dashed circle: chart radius $r_{\rm c}$). Blurring an atom $(n,m)$ by Eq.~\eqref{eq:closure} keeps its order, widens the window and lowers the frequency; labels give the peak amplitude factor. At the handoff $\tau_c$, less than 0.9\% of the energy lies beyond $r_{\rm c}$. The closed form matches FFT heat flow to below $10^{-10}$.}
\label{fig:atoms}
\end{figure}

\paragraph{Heat closure.} Let $H_\tau=e^{\tau\Delta}$ denote heat flow for time $\tau$. On a flat chart, by the Jacobi--Anger expansion and the Gaussian heat kernel,
\begin{equation}
H_\tau\big[g_\rho J_m(\kappa r)e^{\mathrm im\theta}\big]=\tfrac{\rho^2}{\rho'^2}e^{-\tau\kappa^2\rho^2/\rho'^2}\,g_{\rho'}J_m(\kappa' r)e^{\mathrm im\theta},
\label{eq:closure}
\end{equation}
with $\rho'^2=\rho^2+2\tau$ and $\kappa'=\kappa\rho^2/\rho'^2$. A blurred atom is an atom of the same order with a wider window, a lower frequency and a closed-form attenuation (Fig.~\ref{fig:atoms}). Steering plane-wave Gabor atoms instead requires discrete angular sampling, which aliases.

\subsection{Heat-Consistent Forward Process}
\label{sec:forward}

Let $\bar F_0=D\,S^*F_0$ be the band-limited ground truth and $c_0=S^*\bar F_0$. We define the forward process
\begin{equation}
q(c_t\mid c_0)=\mathcal N\!\big(\alpha_t\,\mathcal A_{\tau_t}c_0,\;\sigma_t^2P\big),\qquad \mathcal A_\tau c_0:=S^*H_\tau\bar F_0 .
\label{eq:forward}
\end{equation}
The mean is the token of the heat-blurred texture and moves from fine to coarse as $\tau_t$ grows; the noise is isotropic on the valid range, and with $\alpha_T\approx0$ the prior is a standard Gaussian on $\mathrm{range}(P)$. Up to curvature and handoff errors, the mean is that of blurring diffusion~\cite{bdm} on the surface; the noise is white in token space, and decoded it has covariance $\Phi^{-1}$ (Supp.~\ref{supp:noise}). The difference lies in how the mean is evaluated:
\begin{equation}
\mathcal A_\tau c_0=\begin{cases}\big\langle H_\tau\psi_{i,n,m},Dc_0\big\rangle_{i,n,m} & \tau\le\tau_c,\\ S^*\!\big(\textstyle\sum_i\bar w_i\,[e^{-(\tau-\tau_c)L}u]_i\big) & \tau>\tau_c,\end{cases}
\label{eq:operator}
\end{equation}
followed by the range projection $P$. The first case is local to each chart. The second couples charts through the anchor graph, starting from the local means $u_i=(\mathcal A_{\tau_c}c_0)_{i,0,0}/\langle\psi_{i,0,0},1\rangle$ at the handoff.

\paragraph{Small $\tau$: blurred atoms.} Heat flow is self-adjoint with respect to the area measure, so
\begin{equation}
\big(\mathcal A_\tau c_0\big)_{i,n,m}=\big\langle\psi_{i,n,m},H_\tau\bar F_0\big\rangle=\big\langle H_\tau\psi_{i,n,m},\bar F_0\big\rangle .
\label{eq:adjoint}
\end{equation}
For $\tau\le\tau_c$, $H_\tau\psi$ is the closed form of Eq.~\eqref{eq:closure}, so each token is an inner product with known functions on its own chart. This needs neither a cross-chart matrix nor an eigenbasis, preserves constants exactly, and has relative error $O(K\rho^2)$ at curvature $K$. For charts cut by a barrier, $H_\tau\psi$ is precomputed by Neumann heat flow, which is still self-adjoint, so the two sides of a barrier stay decoupled.

\paragraph{Large $\tau$: anchor-graph heat.} By $\tau_c=2\rho^2$, all surviving frequencies satisfy $\kappa'h\le0.8<\pi$, below the anchor Nyquist rate, so anchor values can carry the signal; continuing with blurred atoms would only widen their support. With $u_i$ the ring-0 token at $\tau_c$ normalized to a local mean, we apply $e^{-(\tau-\tau_c)L}$ by Chebyshev polynomials~\cite{cheb}, with $L=M_a^{-1}(D'-W')$, $W'_{ij}=\frac{m_im_j}{4\pi\delta^2}\exp\!\big(-\bar d_{ij}^2/(4\delta)\big)$ with $\delta=h^2$ for anchor pairs whose unfolding path avoids $\mathcal B$, and masses $m_i=\int\bar w_i\,\mathrm dA$ that conserve the area-weighted mean exactly. Then $L\approx-\Delta$~\cite{belkin}, so $\tau$ is one heat time across $\tau_c$. The result is interpolated with a partition of unity $\bar w_i$ and analyzed. Anchor values carry only the local mean, so order $m\ge1$ structure that remains at $\tau_c$ is dropped. It accounts for 1.6\% of the signal norm at $\tau_c$ on average, and up to 3.4\% on plaid textures (Supp.~\ref{supp:handoff}).

\paragraph{Parameters.} The window $\rho=h/2$ balances conditioning against size. Wider windows lower $B/A$ only slightly (2.76 at $\rho=0.6h$ against 2.86 on a regular grid), while analysis cost grows as $(\rho/h)^2$ (Table~\ref{tab:frame}). The handoff $\tau_c=2\rho^2$ is the earliest time at which all surviving frequencies fall below the anchor Nyquist rate. Earlier handoffs drop more order-$m\ge1$ energy; later ones widen the blurred atoms and slow the operator (Table~\ref{tab:tcsweep}).

\subsection{Gauge-Equivariant Denoiser}
\label{sec:dit}

The denoiser $f_\theta(c_t,t,y)$ is a transformer over the $N$ chart tokens. Its features are organized by order $m$, the irreducible representations of $\mathrm{SO}(2)$. Order-$m$ features are complex vectors that transform like Eq.~\eqref{eq:gauge} under a frame change; order-0 features are invariant.

\textbf{Embedding.} For each order, a complex linear map mixes rings and colour channels. This commutes with Eq.~\eqref{eq:gauge}. Invariant geometric features (curvature, area, distance to $\mathcal B$) and the diffusion time enter through adaptive layer norm on the order-0 path~\cite{dit}.

\textbf{Local attention.} Each token attends to the anchors within its chart radius (about 36). Before aggregation, the order-$m$ features of neighbour $j$ are transported into the frame of $i$ by $e^{\mathrm im\phi_{ij}}$, as in gauge-equivariant mesh networks~\cite{hsn,get}. Attention logits use only invariants (order-0 features, real parts of transported inner products, and a distance bias $-\bar d_{ij}^2/4\tau_{\rm att}(t)$), and nonlinearities gate each order by its norm. Transport inherits an unfolding error of about 1.9$^\circ$ per unit of order~\cite{localhk}, so only orders $|m|\le8$ are exchanged between charts.

\textbf{Global stream.} 256 cluster tokens, obtained by farthest-point sampling on the anchors, pool order-0 features together with their 3D positions and run standard self-attention. Their output is broadcast back as an order-0 shift and as gates. The stream coordinates global appearance. Because it pools 3D positions it is the only Euclidean pathway, but with 256 tokens it is too coarse to tie neighbouring sheets together (Table~\ref{tab:ablation}). Because it acts on invariants only, the network is exactly equivariant to the frame chosen for each chart, with a measured error of $2\times10^{-6}$ in single precision. Transport between charts is only as accurate as the connection angles.

\textbf{Image conditioning.} DINOv2 patch features~\cite{dino} of $y$ are projected onto anchors visible from the estimated camera, with a visibility flag, and enter the order-0 path and the global stream's cross-attention. Image orientation enters as an order-1 input $e^{-\mathrm i\beta_i}$, with $\beta_i$ the angle of the camera's image axis in each visible chart's frame; like a token, it gains $e^{\mathrm i\phi}$ under Eq.~\eqref{eq:gauge}, and its powers give order-$m$ cues.

\subsection{Training, Sampling and Outputs}
\label{sec:training}

\paragraph{Training.} We minimize $\|P f_\theta(c_t,t,y)-c_0\|^2$, which is within a factor $B/A$ of the error in signal space. The loss is weighted per ring with min-SNR~\cite{minsnr}, using $\mathrm{SNR}_n(t)=\alpha_t^2\gamma_n(\tau_t)^2/\sigma_t^2$, where $\gamma_n$ is the closed-form attenuation of ring $n$. Tokens are computed online from ground-truth textures by Eq.~\eqref{eq:adjoint}.

\paragraph{Sampling.} Because the forward mean is linear in $c_0$, DDIM~\cite{ddim} applies with noise restricted to the range. Given $\hat c_0=Pf_\theta$, let $\hat\epsilon=(c_t-\alpha_t\mathcal A_{\tau_t}\hat c_0)/\sigma_t$; we then update $c_{t'}=\alpha_{t'}\mathcal A_{\tau_{t'}}\hat c_0+\sigma_{t'}\hat\epsilon$. We use 50 steps and classifier-free guidance with scale 3.

\paragraph{Example.} Consider one chart at a step with $\tau_{t'}\le\tau_c$. The network returns $133\times C$ coefficients for the chart, which $P$ maps to a valid token by decoding $\hat F=D\hat c_0$ in its neighbourhood with a few conjugate-gradient iterations and re-analysing it. The next mean $\langle H_{\tau_{t'}}\psi_{i,n,m},\hat F\rangle$ is a sum over the chart's quadrature points of $\hat F$ times the closed-form blurred atom of Eq.~\eqref{eq:closure}, and adding $\sigma_{t'}\hat\epsilon$ completes the DDIM step. For $\tau_{t'}>\tau_c$, the blurred-atom sum is replaced by one Chebyshev polynomial on the anchor graph, applied to all charts at once.

\paragraph{Cascade.} For fine detail such as text, a second model with $N_2=65{,}536$ anchors and local attention only handles $\tau<\tau_1$. The split is $\tau_1=2\rho_1^2$, the handoff time of the first model. The fine model receives the fine-level tokens of the first model's decoded output as an extra input.

\paragraph{Outputs and editing.} The decoder evaluates $F$ anywhere, so a sample can be baked into any UV layout, written to vertices, or transferred to another discretization. For repainting, we replace the known region at every step, and a barrier on the mask boundary decouples the two sides.

\section{Experiments}
\label{sec:exp}

\subsection{Setup}
\label{sec:exp:setup}

\paragraph{Training data.} We train on 118k textured Objaverse assets~\cite{objaverse}, deduplicated against all test sets. Half of the training meshes are re-extracted from an SDF by marching cubes, with textures transferred along normal rays, to match the geometry that image-to-3D generators produce; the other half keep their original triangulation. Conditioning images are renders from random views with known cameras.

\paragraph{Benchmarks.} \textbf{GSO-Remesh} uses all 1,030 Google Scanned Objects~\cite{gso}, re-extracted by marching cubes as generators do, with UVs discarded, one input view and 16 held-out views. \textbf{I2M-Wild} contains 500 images generated with GPT-Image-2.5, whose meshes come from the shape stages of Hunyuan3D~2.1~\cite{hunyuan} and TRELLIS.2~\cite{trellis2} and are used as produced; 38\% of them are non-manifold. \textbf{ThinSet-2L} contains 300 held-out Objaverse assets with two-layer parts such as linings, leaves and folded sheets, processed like GSO-Remesh. Pairs that marching cubes fuses into a single sheet cannot be separated by any surface field and are excluded from Leak. Details are in Supp.~\ref{supp:data}.

\paragraph{Metrics.} On GSO-Remesh we report LPIPS~\cite{lpips} over held-out views and two masked variants: LPIPS-occ on pixels invisible from the input view, and LPIPS-thin on points with another sheet within $2h$ along the normal. KID~\cite{kid} ($\times10^3$) is computed against ground-truth renders on GSO-Remesh and against renders of artist-made assets on I2M-Wild. On ThinSet-2L, \emph{Leak} measures how much of the opposite layer's colour appears in the error. For a point $x$ on layer $A$ with opposite point $x'$ on layer $B$,
\begin{equation}
\mathrm{Leak}=\frac{\mathbb E\big[\langle\hat F(x)-F_A(x),\,F_B(x')-F_A(x)\rangle\big]}{\mathbb E\big[\|F_B(x')-F_A(x)\|^2\big]},
\label{eq:leak}
\end{equation}
averaged over both directions. Leak is $0$ without bleeding and $1$ when a layer takes on the other layer's colour. Pref.\ is the share of votes in the blind user study (Sec.~\ref{sec:exp:preference}).

\paragraph{Baselines.} We compare with TEXGen~\cite{texgen} (UV space), TRELLIS.2~\cite{trellis2} (sparse-voxel latents), Hunyuan3D-Paint~2.1~\cite{hunyuan} (multi-view generation and baking) and NaTex~\cite{natex} (colour point latents), using released checkpoints with the same image and camera. LaFiTe~\cite{lafite} has no public checkpoint, so we retrain it on our data with our budget. GOATex~\cite{goatex}, which handles occluded layers explicitly, has no public code; we reimplement it on top of the Hunyuan3D-Paint multi-view generator (Supp.~\ref{supp:baselines}).

\paragraph{Implementation.} HeatGen uses $N=16{,}384$ anchors, 24 blocks and about 420M parameters, and trains for 160 H100-days. The cascade adds a local-only model with 65,536 anchors. Controlled studies (Fig.~\ref{fig:forward} and Table~\ref{tab:ablation}) run at a smaller \emph{M-scale} ($N=4{,}096$, 1/4 of the iterations) with three seeds. Details are in Supp.~\ref{supp:impl}. We will release code, models and ThinSet-2L.

\begin{table*}[t]
\centering
\small
\setlength{\tabcolsep}{7pt}
\begin{tabular}{l cccc cc c}
\toprule
& \multicolumn{4}{c}{GSO-Remesh} & \multicolumn{2}{c}{I2M-Wild} & ThinSet-2L \\
\cmidrule(lr){2-5}\cmidrule(lr){6-7}
Method & LPIPS$\downarrow$ & LPIPS-occ$\downarrow$ & LPIPS-thin$\downarrow$ & KID$\times10^3$$\downarrow$ & KID$\times10^3$$\downarrow$ & Pref.\ (\%)$\uparrow$ & Leak$\downarrow$ \\
\midrule
TEXGen~\cite{texgen} & 0.171 & 0.214 & 0.238 & 14.9 & 18.2 & 4.1 & 0.09 \\
TRELLIS.2~\cite{trellis2} & 0.158 & 0.196 & 0.231 & 12.6 & 14.0 & 7.3 & 0.29 \\
Hunyuan3D-Paint 2.1~\cite{hunyuan} & \underline{0.149} & 0.193 & 0.224 & \underline{11.2} & \textbf{12.4} & \underline{17.8} & 0.12 \\
NaTex~\cite{natex} & 0.152 & 0.188 & 0.227 & 11.9 & 12.9 & 10.6 & 0.25 \\
LaFiTe~\cite{lafite} & 0.153 & 0.190 & 0.219 & 11.8 & 12.8 & 9.4 & 0.13 \\
GOATex~\cite{goatex} & 0.155 & \underline{0.185} & \underline{0.217} & 11.9 & 12.9 & 13.2 & \underline{0.07} \\
\midrule
HeatGen (Ours) & \textbf{0.135} & \textbf{0.171} & \textbf{0.198} & \textbf{10.5} & \underline{12.7} & \textbf{37.6} & \textbf{0.036} \\
\bottomrule
\end{tabular}
\caption{Quantitative comparison on image-to-mesh texturing. We report LPIPS on all held-out views and on occluded and thin regions of GSO-Remesh, KID on GSO-Remesh and on the generated meshes of I2M-Wild, the share of user votes, and leakage between layers on ThinSet-2L. HeatGen is the best or close to the best on overall metrics and clearly better on occluded and thin regions. HeatGen metrics excluding Pref. are means over three training seeds. Best and second-best results are in bold and underlined.}
\label{tab:i2m}
\end{table*}

\begin{figure*}[t]
\centering
\input{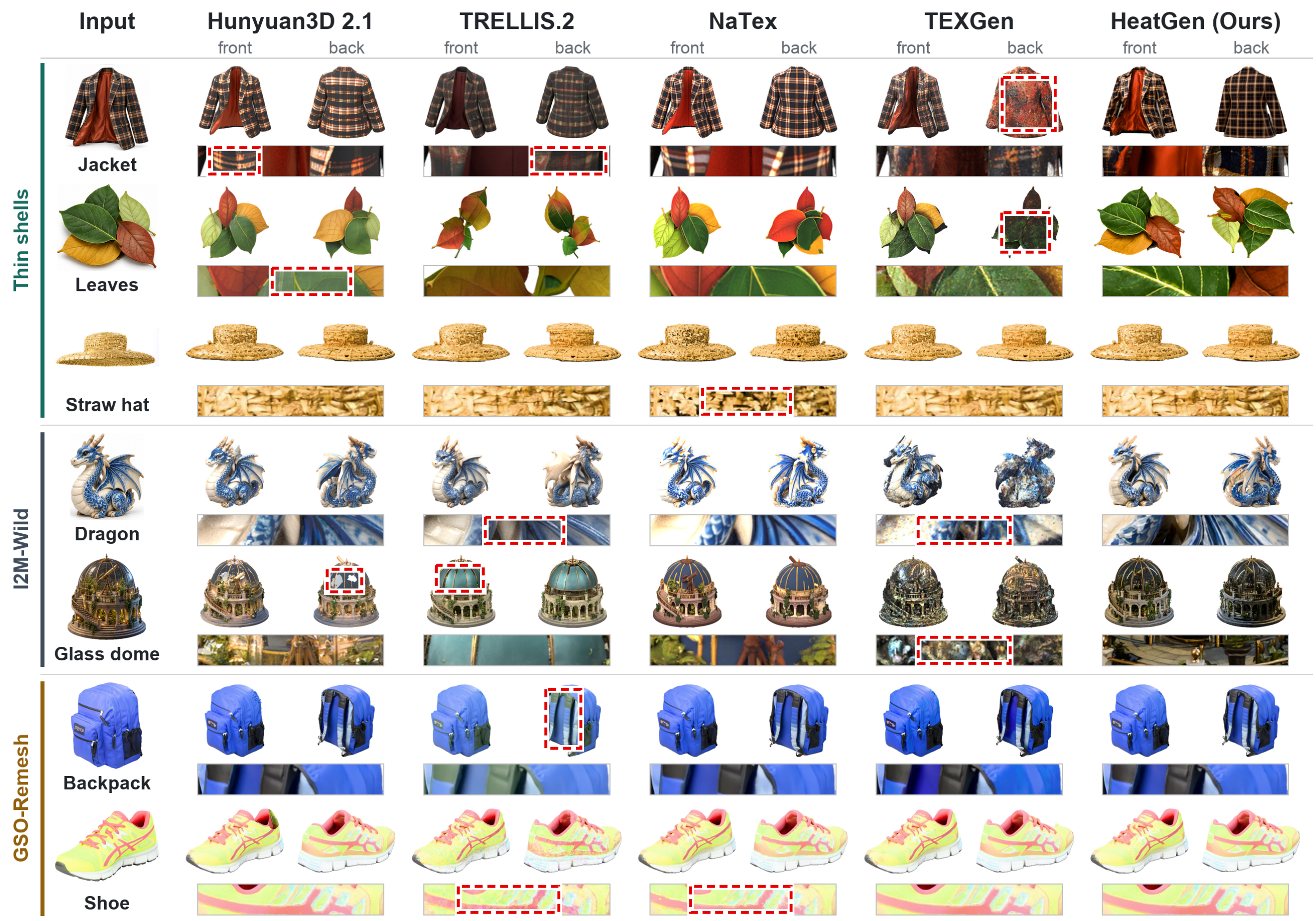}
\caption{Examples of qualitative comparison, grouped into thin-shell objects, other objects of I2M-Wild, and scans of GSO-Remesh. The methods shown texture the same mesh from the input image. For each method we show a view near the input (front), the opposite view (back), and a close-up of a thin or contact region (strip below). Red dashed boxes mark visible artifacts. The multi-view baseline must inpaint the back from the front views, and the native 3D latents mix the colours of nearby sheets, such as the jacket lining and shell or overlapping leaves. HeatGen completes the unseen side consistently with the input and keeps the two sides of thin shells distinct.}
\label{fig:qual}
\end{figure*}

\subsection{Image-to-Mesh Texturing}
\label{sec:exp:i2m}

Table~\ref{tab:i2m} and Fig.~\ref{fig:qual} compare complete systems. HeatGen has the best overall LPIPS and KID on GSO-Remesh, with KID 10.5\ versus 11.2 for Hunyuan3D-Paint~2.1. On I2M-Wild, HeatGen ranks second in KID (\underline{12.7} versus \textbf{12.4}), while receiving the largest preference share. A second, finer level (the cascade) improves LPIPS and layer separation at about twice the compute; its GSO-Remesh KID is 10.3\ (Supp.~\ref{supp:more}).

The gains are largest where projection and Euclidean latents have difficulty. On pixels unseen by the input view, HeatGen improves LPIPS from 0.185 for the occlusion-aware GOATex to 0.171, and on thin and contact pixels from 0.217 to 0.198, at every thin-mask threshold (Supp.~\ref{supp:metrics}). On ThinSet-2L, every baseline transfers at least 0.07 of the opposite layer's colour, while HeatGen transfers 0.036. Such regions are common: 64\% of I2M-Wild meshes contain thin or touching parts. HeatGen takes 37\,s per asset, between the native systems and Hunyuan3D-Paint.

\paragraph{Effect of the representation.} Data and conditioning also differ between systems, so we train other surface states with exactly our data, budget and pixel-aligned conditioning (Supp.~\ref{supp:unique}). A point latent with geodesic attention and decoding reaches LPIPS-occ 0.186 and Leak 0.061, and enlarging it or adding learned frames changes little; chart tokens lower LPIPS-occ by 0.015 (bootstrap interval 0.011--0.019) and reduce Leak to 0.036. The difference is larger where the mesh changes: across five triangulations of the same shape, chart-token textures vary by 0.031 LPIPS, against 0.047 for the point latent.

\subsection{Layer Separation}
\label{sec:exp:thin}

\begin{table}[t]
\centering
\small
\resizebox{\linewidth}{!}{\begin{tabular}{l cc ccc}
\toprule
& \multicolumn{2}{c}{Size (M)} & & & \\
\cmidrule(lr){2-3}
Representation & Latent & Decoder$\downarrow$ & PSNR (dB)$\uparrow$ & PSNR-thin (dB)$\uparrow$ & Leak$\downarrow$ \\
\midrule
\multicolumn{6}{l}{\emph{Explicit storage (reference)}} \\
UV atlas, $1024^2$ & 3.1 & 0 & 31.4 & 30.6 & 0.00 \\
Vertex colours & 0.6 & 0 & 27.9 & 26.1 & 0.02 \\
\midrule
\multicolumn{6}{l}{\emph{Learned latents}} \\
NaTex~\cite{natex} & 0.4 & 300 & 30.2 & 24.7 & 0.26 \\
TRELLIS.2~\cite{trellis2} & 0.3 & 474 & 29.6 & 23.9 & 0.31 \\
LaFiTe~\cite{lafite} & 2.0 & 280 & 30.6 & 26.1 & 0.13 \\
Colour point latent & 6.5 & 120 & 30.4 & 25.0 & 0.17 \\
\midrule
\multicolumn{6}{l}{\emph{Ours}} \\
Chart tokens, $N{=}16$k & 6.5 & \textbf{0} & \underline{31.3} & \underline{30.2} & \textbf{0.01} \\
Chart tokens, $N{=}64$k & 26.1 & \textbf{0} & \textbf{33.6} & \textbf{32.9} & \textbf{0.01} \\
\bottomrule
\end{tabular}}
\caption{Fidelity of texture representations. Ground-truth textures are encoded and decoded without generation. Size lists the per-asset latent and the decoder parameters; explicit storage is shown for reference and not ranked.}
\label{tab:repr}
\end{table}

Table~\ref{tab:repr} evaluates the representations alone by encoding and decoding ground-truth textures. Since a learned latent cannot be read without its decoder, we report the size of both. A UV atlas and vertex colours store every surface point separately and therefore exhibit nearly no leakage, but they need an atlas or follow the triangulation. The learned latents rely on decoders with 120--474M parameters and still lose 4.5--5.7\,dB on thin pixels and leak 0.13--0.31, because opposite layers share latent cells. Chart tokens make the opposite trade-off: their latent is larger, because the analysis is redundant rather than learned, but they are decoded by a fixed linear operator without any learned parameters. Chart tokens have the best overall PSNR among the latents, lose only 1.1\,dB on thin pixels and keep Leak at 0.01. Their resolution is set by the anchors rather than the mesh: at $16$k anchors they are within 0.4\,dB of a $1024^2$ atlas on thin pixels, and at $64$k they exceed it, without an atlas or seams.

Separation weakens where two layers meet at a hem or rim, because charts there span both sides. HeatGen's Leak is 0.012 for separate layers and 0.051 for joined ones, and even close to a junction it stays below every baseline (Supp.~\ref{supp:leak}). On the hand-designed garment layers of CLOTH4D~\cite{cloth4d}, used as authored, HeatGen leaks 0.021 against 0.06 for GOATex.

\subsection{Forward Process Analysis}
\label{sec:exp:forward}

\begin{figure}[t]
\centering
\begingroup
\definecolor{fwdours}{RGB}{37,99,235}
\definecolor{fwdbase}{RGB}{100,116,139}
\definecolor{fwdink}{RGB}{51,65,85}
\newcommand{\fwdbubble}[6]{%
  \pgfmathsetmacro{\fwdr}{0.8*sqrt(#3)}%
  \def\fwdpos{}%
  \if#6w\edef\fwdpos{anchor=west,xshift=\fwdr pt+2.5pt}\fi
  \if#6n\edef\fwdpos{anchor=north,yshift=-\fwdr pt-1.5pt}\fi
  \if#6s\edef\fwdpos{anchor=south,yshift=\fwdr pt+1.5pt}\fi
  \edef\fwdcmd{\noexpand\addplot[only marks,mark=*,mark size=\fwdr pt,draw=#4!80!black,line width=0.5pt,fill=#4,fill opacity=0.72,draw opacity=1] coordinates {(#1,#2)};\noexpand\node[font=\noexpand\footnotesize,text=#4!70!black,inner sep=1pt,\fwdpos] at (axis cs:#1,#2) {\noexpand#5};}%
  \fwdcmd}
\begin{tikzpicture}
\begin{axis}[
  width=\linewidth, height=0.72\linewidth,
  xmode=log, log basis x=10,
  xmin=3.2, xmax=70, ymin=14.6, ymax=17.3,
  xtick={5,10,20,40}, xticklabels={5,10,20,40},
  ytick={15,16,17},
  xlabel={Sampling time to reach the Gaussian baseline's KID (s)},
  ylabel={KID$\times10^3$},
  label style={font=\footnotesize, text=fwdink}, tick label style={font=\footnotesize, text=fwdink},
  axis x line*=bottom, axis y line*=left,
  axis line style={fwdink!45, line width=0.5pt}, tick style={fwdink!45, line width=0.5pt},
  major tick length=2.5pt, tick align=outside,
  grid=major, grid style={fwdink!10, line width=0.4pt},
  clip=false]
\fill[fwdours, fill opacity=0.16] (axis cs:4.9,15.1) circle[radius=0.8*sqrt(55)*1pt+3pt];
\fwdbubble{11.2}{16.8}{100}{fwdbase}{Gaussian (VP)}{w}
\fwdbubble{9.0}{16.0}{74}{fwdbase}{Per-ring schedule}{w}
\fwdbubble{6.8}{15.4}{61}{fwdbase}{Per-ring + graph heat}{w}
\fwdbubble{36.2}{15.3}{58}{fwdbase}{Mesh heat}{s}
\fwdbubble{4.9}{15.1}{55}{fwdours}{\textbf{Ours}}{n}
\pgfmathsetmacro{\fwdrs}{0.8*sqrt(50)}\pgfmathsetmacro{\fwdrl}{0.8*sqrt(100)}
\path (rel axis cs:0,1) coordinate (fwdL);
\draw[rounded corners=2pt, fill=white, draw=fwdink!22, line width=0.4pt]
  ([xshift=4pt,yshift=-3pt]fwdL) rectangle ([xshift=56pt,yshift=-43pt]fwdL);
\node[anchor=north, font=\scriptsize, text=fwdink, inner sep=0pt] at ([xshift=30pt,yshift=-6pt]fwdL) {Iters};
\draw[fwdbase!80!black, fill=fwdbase!30, line width=0.4pt] ([xshift=17pt,yshift=-31pt+\fwdrs pt]fwdL) circle[radius=\fwdrs pt];
\draw[fwdbase!80!black, fill=fwdbase!30, line width=0.4pt] ([xshift=41pt,yshift=-31pt+\fwdrl pt]fwdL) circle[radius=\fwdrl pt];
\node[anchor=north, font=\scriptsize, text=fwdink, inner sep=0pt] at ([xshift=17pt,yshift=-33pt]fwdL) {50\%};
\node[anchor=north, font=\scriptsize, text=fwdink, inner sep=0pt] at ([xshift=41pt,yshift=-33pt]fwdL) {100\%};
\draw[-{Stealth[length=4.5pt,width=3.5pt]}, fwdink!40, line width=0.7pt] (axis cs:21,15.2) -- (axis cs:12.5,14.84)
  node[pos=0.45, sloped, below, font=\scriptsize\itshape, text=fwdink!60, inner sep=2pt] {better};
\end{axis}
\end{tikzpicture}
\endgroup
\caption{Comparison of forward processes at M-scale. Time is the sampling time needed to reach the final KID of the Gaussian baseline, and bubble area shows Iters, the share of training iterations needed to reach that KID. All variants share the tokens, network and tuning budget.}
\label{fig:forward}
\end{figure}
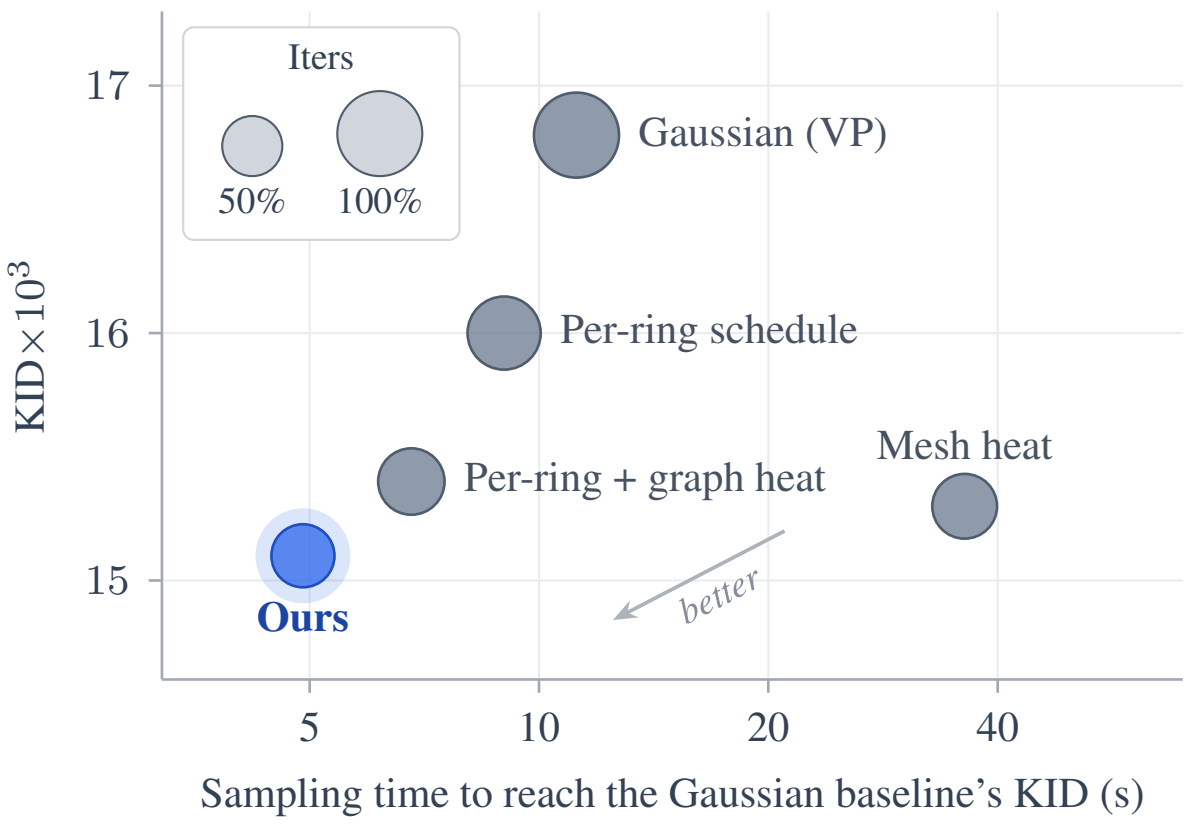

Fig.~\ref{fig:forward} varies only the forward process. The Gaussian baseline adds noise with one variance-preserving schedule for all frequencies. The per-ring schedule tunes a separate schedule for each ring, the strongest schedule that is diagonal in the token basis. Per-ring + graph heat keeps this schedule at fine scales and uses our anchor-graph heat at coarse scales. Mesh heat computes exact heat flow at mesh resolution with Krylov methods~\cite{expmv}. All variants share the tokens, network and tuning budget. Numerical values and the log-spectral distance are in Supp.~\ref{supp:replication}.

Our process improves on the per-ring schedule from 16.0 to 15.1 KID and reaches the final quality of the Gaussian baseline with 55\% of the iterations. Coupling charts at coarse scales gives most of the gain, and the closed-form blur at fine scales the rest. Because coarse scales are settled first, sampling also converges sooner: our process reaches the Gaussian baseline's final KID with 20 instead of 50 DDIM steps, in 4.9\,s instead of 11.2\,s. Our process matches exact mesh heat in quality (15.1 against 15.3 KID) at a fraction of its cost: mesh heat must be evaluated at mesh resolution twice per sampling step, which makes 50-step sampling 5.9$\times$ slower at M-scale and 3.0$\times$ at full scale. At full scale, with three training seeds per variant, the ordering is the same (Supp.~\ref{supp:replication}).

\subsection{Ablation Study}
\label{sec:exp:ablation}

\begin{table}[t]
\centering
\small
\setlength{\tabcolsep}{3.2pt}
\resizebox{\linewidth}{!}{\begin{tabular}{l cccc}
\toprule
Variant & KID$\downarrow$ & LPIPS-occ$\downarrow$ & LPIPS-thin$\downarrow$ & Leak$\downarrow$ \\
\midrule
Full model & \textbf{15.1} & \textbf{0.191} & \textbf{0.214} & \textbf{0.041} \\
\midrule
\multicolumn{5}{l}{\emph{State}} \\
Fourier--Bessel $\to$ Gabor atoms & 16.0 & 0.197 & 0.221 & 0.046 \\
Chart tokens $\to$ vertex colours & 16.6 & 0.199 & 0.224 & 0.049 \\
Chart tokens $\to$ point latent (Euclidean) & 15.4 & 0.199 & 0.233 & 0.187 \\
\midrule
\multicolumn{5}{l}{\emph{Network}} \\
w/o gauge equivariance & 16.3 & 0.198 & 0.223 & 0.044 \\
Geodesic $\to$ Euclidean attention & 15.2 & 0.192 & 0.229 & 0.168 \\
w/o global stream & 16.7 & 0.201 & 0.218 & 0.044 \\
\bottomrule
\end{tabular}}
\caption{Ablation study at M-scale. Each row replaces ($\to$) or removes (w/o) one component of HeatGen, with data, budget and image conditioning fixed. Results are averaged over three seeds and are not comparable with Table~\ref{tab:i2m}.}
\label{tab:ablation}
\end{table}

Table~\ref{tab:ablation} removes one component at a time, keeping data, budget and image conditioning fixed; further variants are in Supp.~\ref{supp:ablation}. The full model is best on every metric. Plane-wave Gabor atoms must be steered by discrete directions, which aliases and raises KID. Vertex colours are limited by mesh resolution, and a Euclidean point latent comes close in KID but leaks. Without gauge equivariance, that is, with one fixed tangent field per mesh, every metric degrades. Euclidean attention keeps KID but reintroduces leakage, so geodesic neighbourhoods matter for separation, not overall quality. Without the global stream, global coherence suffers and leakage rises slightly.

\subsection{User Preference Study}
\label{sec:exp:preference}

We assess perceptual quality on I2M-Wild through a blind study with 48 participants, including 12 3D artists. Each participant judges 40 assets, giving 1,920 votes, and picks the texture that best matches the input image and looks most plausible among turntable videos of all seven methods (protocol in Supp.~\ref{supp:study}).

HeatGen receives 37.6\% of the votes, about twice the share of the next method, Hunyuan3D-Paint (17.8\%; Table~\ref{tab:i2m}). A second study focuses on thin and contact regions: 24 further participants each judge 30 assets using zoomed turntables. HeatGen's share rises to 45.1\%, compared with 17.4\% for the strongest baseline, GOATex. This agrees with the lower LPIPS-thin and Leak.

\section{Limitations}
\label{sec:limitations}

\paragraph{Dependence on input topology.} Layer separation follows the supplied surface connectivity. If reconstruction fuses two layers into one sheet, a surface colour field cannot recover their distinct appearances. At a genuine hem or rim, charts may span both sides near the junction (Sec.~\ref{sec:exp:thin}). Geometry refinement could complement HeatGen by restoring missing layers before texturing; local barriers can restrict transport when a seam is known.

\paragraph{Fine detail and compute.} Anchor spacing sets the finest texture band. Small print and sharp patterns benefit from the cascade, whose denser anchors and smaller charts increase memory and sampling cost (Supp.~\ref{supp:more}). Adaptive anchors could concentrate this cost on detailed regions.

\section{Conclusion}
\label{sec:conclusion}

HeatGen diffuses chart tokens, the analysis coefficients of steerable, heat-closed atoms on geodesic charts. Analytically blurred atoms and anchor-graph heat give an efficient surface-consistent forward process without an eigenbasis, while a gauge-equivariant transformer denoises the tokens. The resulting textures match or improve on strong systems overall, improve occluded and thin regions, and stay consistent across discretizations.

{
  \small
  \bibliographystyle{ieeenat_fullname}
  \bibliography{references}
}

\clearpage
\appendix
\renewcommand{\thetable}{S\arabic{table}}
\renewcommand{\thefigure}{S\arabic{figure}}
\renewcommand{\thealgorithm}{S\arabic{algorithm}}
\setcounter{table}{0}
\setcounter{figure}{0}
\setcounter{algorithm}{0}
\section*{Supplementary Material}
\section{Implementation Details}
\label{supp:impl}

This section gives the details needed to reproduce HeatGen: charts, atoms and the forward operator (\ref{supp:charts}--\ref{supp:schedule}), the network (\ref{supp:arch}), training and sampling (\ref{supp:trainsample}), the data pipeline (\ref{supp:pipeline}) and pseudocode (\ref{supp:algos}). Table~\ref{tab:hyper} summarizes all hyperparameters.

\begin{table}[!h]
\centering
\small
\resizebox{\linewidth}{!}{\begin{tabular}{l l}
\toprule
Hyperparameter & Value \\
\midrule
\multicolumn{2}{l}{\emph{Representation}} \\
Anchors $N$ (full / M-scale / cascade) & 16,384 / 4,096 / 65,536 \\
Chart radius $r_{\rm c}$ & $3.4h$ \\
Window $\rho$ & $h/2$ \\
Rings / orders & $n=0,\dots,6$ / $|m|\le3n$ \\
Atoms per chart & 133 \\
\midrule
\multicolumn{2}{l}{\emph{Forward process}} \\
Heat time & log-uniform, $\tau_{\min}=0.01\rho^2$ \\
Handoff $\tau_c$ & $2\rho^2$ \\
Anchor-graph heat & Chebyshev, degree 30 \\
Noise schedule & cosine, variance preserving \\
\midrule
\multicolumn{2}{l}{\emph{Network}} \\
Blocks / heads & 24 / 12 \\
Width ($m{=}0$ / $1$--$4$ / $5$--$8$ / ${>}8$) & 768 / 256 / 128 / 32 \\
Neighbours / exchanged orders & ${\approx}36$ / $|m|\le8$ \\
Global tokens & 256 \\
Image encoder & DINOv2-L \\
Parameters & 420M \\
\midrule
\multicolumn{2}{l}{\emph{Training}} \\
Optimizer & AdamW, $\beta=(0.9,0.99)$, weight decay 0.01 \\
Learning rate & $10^{-4}$, 5k warm-up, cosine decay \\
Batch / iterations & 64 / 400k (cascade: 200k) \\
EMA / condition dropout & 0.9999 / 10\% \\
Loss weighting & ring-wise min-SNR, $\gamma=5$ \\
Precision & bfloat16; float32 for CG and Chebyshev \\
Compute & 160 H100-days \\
\midrule
\multicolumn{2}{l}{\emph{Sampling}} \\
Sampler / steps & DDIM ($\eta=0$) / 50 \\
Guidance scale & 3 \\
\bottomrule
\end{tabular}}
\caption{Hyperparameters of HeatGen.}
\label{tab:hyper}
\end{table}

\subsection{Charts and Atoms}
\label{supp:charts}
\paragraph{Proxy surface.} Non-manifold inputs are replaced by their tufted cover~\cite{robustlap}. This duplicates faces around non-manifold edges so that every edge has at most two incident faces, while keeping the geometry. At non-manifold vertices, hinge unfolding passes between faces only across shared edges, so each sheet that meets at the vertex is unfolded on its own. All later steps run on this cover, and outputs are mapped back to the input faces.

\paragraph{Anchors.} We sample $N$ anchors by farthest-point sampling on the surface, using geodesic distances from a fast-marching front. We then run five Lloyd iterations with geodesic Voronoi cells. Lloyd relaxation lowers the frame bound ratio from 3.41 to a median of 2.94 (Table~\ref{tab:checks}).

\paragraph{Unfolding.} Each anchor is unfolded with the batched hinge unfolding of~\cite{localhk}. Faces are processed once, in rings of increasing distance from $p_i$, and each new face is attached to an unfolded neighbour by the rigid rotation about their shared edge. Faces beyond $r_{\rm c}$ or across a barrier edge are skipped. At a non-manifold edge of the input, the tufted cover offers one continuation per incident sheet, so a chart never folds two sheets onto one plane.
Let $\mathcal C_i$ be the unfolded chart and $U_i(q)\in\mathbb R^2$ its planar coordinates, with $U_i(p_i)=0$. Area quadrature gives nodes $Q_i$ and area weights $\{a_q\}$, and every node receives polar coordinates $(r_i(q),\theta_i(q))$. The ordered product of hinge rotations $R_e$ across edges along the unfolding path $\pi_i(f)$ gives the frame $R_i(f)$ at face $f$.

\paragraph{Connection angles.} For overlapping charts $i,j$, let $\Delta_{ij}(f)$ be the signed relative angle between $R_i(f)$ and $R_j(f)$ on a shared face, in the transport convention of Sec.~\ref{sec:dit}. The median of these angles gives $\phi_{ij}$. It is antisymmetrized as $\phi_{ij}\leftarrow\phi_{ij}-\tfrac12\mathrm{wrap}_{(-\pi,\pi]}(\phi_{ij}+\phi_{ji})$.

\paragraph{Atoms and quadrature.} Rings $\kappa_n\rho=2n$ for $n=0,\dots,6$ with orders $|m|\le3n$ give 133 real-packed atoms per chart and channel, normalized in the flat chart metric. Inner products use about 64 area-weighted quadrature points per $h^2$, enough to resolve $k_{\max}\approx25/h$. The table of quadrature coordinates is the main memory cost of preprocessing (Supp.~\ref{supp:more}).

\paragraph{Decoder.} We apply $\Phi^{-1}$ by conjugate gradients on the anchor-weighted quadrature points, with relative tolerance $10^{-4}$. Given the frame bound, this takes 6--9 iterations. The range projection $P=S^*\Phi^{-1}S$ reuses the same solve.

\subsection{Forward Operator and Schedule}
\label{supp:schedule}
Heat time is log-uniform, $\tau_t=\tau_{\min}(\tau_{\max}/\tau_{\min})^{t/T}$, with $\tau_{\min}=0.01\rho^2$ and $\tau_{\max}$ the squared mean geodesic radius of the mesh, so every octave of scale receives the same number of steps. At $N=16{,}384$, 41\% of the steps have $\tau_t\le\tau_c$; at $N=4{,}096$, 45\%. The noise levels $(\alpha_t,\sigma_t)$ follow a cosine variance-preserving schedule, tuned within the same 20-trial budget as every variant in Fig.~\ref{fig:forward}.

For $\tau\le\tau_c$, blurred atoms are evaluated in closed form at the same quadrature points as the analysis. Charts cut by a barrier or a mesh boundary store Neumann heat responses at 16 heat times, computed once on the chart patch, and interpolate between them. We denote the chartwise Neumann heat operator by $H^{\mathrm N,i}_\tau$ and its cached responses by $\Psi^{(\ell)}_{i,n,m}$. For $\tau>\tau_c$, the anchor graph uses $\delta=h^2$, a connection threshold of twice the covering radius, and degree-30 Chebyshev polynomials for $e^{-(\tau-\tau_c)L}$. At the handoff, $u_i=(\mathcal A_{\tau_c}c_0)_{i,0,0}/\langle\psi_{i,0,0},1\rangle$ is the window-weighted local mean at $\tau_c$.

\subsection{Network Architecture}
\label{supp:arch}
The network has 24 blocks. Each block applies an order-wise local attention, an order-wise gated MLP, a global-stream update with image cross-attention, and a broadcast from the global stream back to the tokens. Complex order-$m$ features are stored as pairs of real channels. All maps on order-$m$ features commute with phase rotations, so the network is equivariant to per-chart frame changes (Supp.~\ref{supp:theory}).

\subsection{Training, Sampling and the Cascade}
\label{supp:trainsample}
Optimization and sampling settings are listed in Table~\ref{tab:hyper}.

\paragraph{Loss.} The loss is $\sum_n w_n(t)\|(Pf_\theta)_n-c_{0,n}\|^2$, with ring-wise min-SNR weights $w_n=\min(\mathrm{SNR}_n,5)/\mathrm{SNR}_n$ and $\mathrm{SNR}_n(t)=\alpha_t^2\gamma_n(\tau_t)^2/\sigma_t^2$. Because $A\le\Phi\le B$, the token error bounds the signal error within a factor $B/A$.

\paragraph{Cascade.} The fine model uses $N_2$ anchors and handles $\tau<\tau_1=2\rho_1^2$, the handoff time of the coarse model. It has no global stream. It receives the fine-level analysis of the decoded coarse sample as an extra input, through the same order-wise embedding. At inference it starts from the coarse sample, noised to $\tau_1$.

\paragraph{Guidance.} Conditional and unconditional predictions share one network, and the null condition replaces image features with a learned token. The guided combination of the two predictions is projected by $P$ (Alg.~\ref{alg:sample}). Supp.~\ref{supp:studies} varies the number of steps.

\subsection{Data Pipeline}
\label{supp:pipeline}
\emph{Filtering.} From Objaverse we keep assets with a texture map of at least $512^2$ texels, which leaves 163k. We then remove assets whose identifier or geometry hash matches any test asset (2,317 removed). We keep 118k after a final quality filter on rendered views. \emph{Re-extraction.} For half of the assets, the mesh is re-extracted from its SDF at $512^3$ by marching cubes. The texture is transferred along normal rays, and points whose ray hits both sides of a fused region are dropped from the loss. \emph{Online tokens.} Tokens are computed on the fly from the texture, so no tokens are stored. Charts are rebuilt by the data loaders, which keeps up with training at 64 assets per step on 32 CPU cores per GPU. \emph{Views.} Each training sample uses one random view with elevation in $[-15^\circ,45^\circ]$ and a known camera.

\subsection{Pseudocode}
\label{supp:algos}
In Alg.~\ref{alg:charts}, $\mathcal E_{\mathcal B}$ contains the overlapping anchor pairs admitted by the graph threshold whose unfolding paths avoid $\mathcal B$; $\mathrm{wrap}$ maps angles to $(-\pi,\pi]$.

\begin{algorithm}[tb]
\caption{Chart construction (once per mesh)}
\label{alg:charts}
\small
\begin{algorithmic}[1]
\State $\mathcal M\gets\operatorname{TuftedCover}(\mathcal M)$ if non-manifold
\State $\{p_i\}_{i=1}^N\gets\operatorname{Lloyd}^{5}(\operatorname{FPS}_N(\mathcal M))$
\State $h\gets\operatorname{MeanSpacing}(\{p_i\})$; $r_{\rm c}\gets3.4h$, $\delta\gets h^2$
\For{$i=1,\dots,N$ (batched)}
  \State $(\mathcal C_i,U_i)\gets\operatorname{HingeUnfold}(\mathcal M,p_i,r_{\rm c},\mathcal B)$
  \State $(Q_i,\{a_q\})\gets\operatorname{AreaQuadrature}(\mathcal C_i)$
  \State $(r_i(q),\theta_i(q))\gets(\|U_i(q)\|,\arg U_i(q))$, $q\in Q_i$
  \State $R_i(f)\gets\prod_{e\in\pi_i(f)}^{\longrightarrow}R_e$ \Comment{ordered hinge rotations}
\EndFor
\State $\widetilde\phi_{ij}\gets\operatorname{median}_{f\in\mathcal C_i\cap\mathcal C_j}\Delta_{ij}(f)$
\State $\phi_{ij}\gets\widetilde\phi_{ij}-\tfrac12\operatorname{wrap}(\widetilde\phi_{ij}+\widetilde\phi_{ji})$
\State $m_i\gets\int_{\mathcal M}\bar w_i\,\mathrm dA$; $M_a\gets\operatorname{diag}(m)$
\State $W'_{ij}\gets \frac{m_im_j}{4\pi\delta^2}\exp\!\big(-\bar d_{ij}^2/(4\delta)\big)\mathbf1_{(i,j)\in\mathcal E_{\mathcal B}}$
\State $D'_{ii}\gets\sum_j W'_{ij}$; $L\gets M_a^{-1}(D'-W')$
\For{each barrier- or boundary-cut chart $i$}
  \State $\Psi^{(\ell)}_{i,n,m}\gets H^{\mathrm N,i}_{\tau_\ell}\psi_{i,n,m}$, $\ell=1,\dots,16$
\EndFor
\end{algorithmic}
\end{algorithm}

\begin{algorithm}[tb]
\caption{One training step}
\label{alg:train}
\small
\begin{algorithmic}[1]
\State sample asset $(\mathcal M,F_0,y)$ and $t\sim\mathcal U\{1,\dots,T\}$; $\tau\gets\tau_t$
\State $c_0\gets P\,S^*F_0$ \Comment{analysis, then projection}
\If{$\tau\le\tau_c$}
  \State $\mu\gets\langle H_\tau\psi_{i,n,m},D c_0\rangle$ for all $(i,n,m)$ \Comment{Eq.~\eqref{eq:closure}, \eqref{eq:adjoint}}
\Else
  \State $u\gets$ ring-0 tokens of $\mathcal A_{\tau_c}c_0$ divided by atom masses
  \State $u\gets e^{-(\tau-\tau_c)L}u$ by Chebyshev; $\mu\gets S^*\!\big(\sum_i\bar w_iu_i\big)$
\EndIf
\State $c_t\gets\alpha_t\mu+\sigma_tP\epsilon$, \ $\epsilon\sim\mathcal N(0,I)$
\State $\hat c_0\gets P f_\theta(c_t,t,y)$; \ minimize $\sum_n w_n(t)\|\hat c_{0,n}-c_{0,n}\|^2$
\end{algorithmic}
\end{algorithm}

\begin{algorithm}[tb]
\caption{Sampling}
\label{alg:sample}
\small
\begin{algorithmic}[1]
\State build charts (Alg.~\ref{alg:charts}); project image features to visible anchors
\State $c_T\gets P\epsilon$
\For{$t=T,\dots,1$ (50 DDIM steps)}
  \State $\hat c_0\gets P\big[(1+w)f_\theta(c_t,t,y)-wf_\theta(c_t,t,\varnothing)\big]$
  \State $\hat\epsilon\gets(c_t-\alpha_t\mathcal A_{\tau_t}\hat c_0)/\sigma_t$
  \State $c_{t-1}\gets\alpha_{t-1}\mathcal A_{\tau_{t-1}}\hat c_0+\sigma_{t-1}\hat\epsilon$ \Comment{$\mathcal A$ as in Alg.~\ref{alg:train}}
\EndFor
\State \Return $\hat F=D\hat c_0$, evaluated at UV texels, vertices or points
\end{algorithmic}
\end{algorithm}

\section{Extended Related Work}
\label{supp:related}

Table~\ref{tab:position} compares HeatGen with the closest texture generators and generative models on surfaces. View-based and UV-space methods need an atlas or baking; native 3D latents are atlas-free but aggregate in 3D; intrinsic models aggregate along the surface but store their state on vertices or points and are mostly trained per shape or category.

\begin{table}[tb]
\centering
\small
\resizebox{\linewidth}{!}{\begin{tabular}{l ccccc}
\toprule
Method & \twoline{Atlas-}{free} & \twoline{Surface}{neighbourhood} & \twoline{Mesh-indep.}{resolution} & \twoline{Cross-}{shape} & \twoline{Heat-based}{forward} \\
\midrule
TEXGen~\cite{texgen} & \xmark & \xmark & \cmark & \cmark & \xmark \\
Hunyuan3D-Paint~\cite{hunyuan} & \xmark & \xmark & \cmark & \cmark & \xmark \\
GOATex~\cite{goatex} & \xmark & \xmark & \cmark & -- & \xmark \\
TRELLIS.2~\cite{trellis2} & \cmark & \xmark & \cmark & \cmark & \xmark \\
NaTex~\cite{natex} & \cmark & \xmark & \cmark & \cmark & \xmark \\
LaFiTe~\cite{lafite}, TEXTRIX~\cite{textrix} & \cmark & \xmark & \cmark & \cmark & \xmark \\
MDF~\cite{mdf} & \cmark & \cmark & \xmark & \xmark & \xmark \\
UV3-TeD~\cite{uv3ted} & \cmark & \cmark & \cmark & \cmark & \xmark \\
DoubleDiffusion~\cite{double} & \cmark & \cmark & \xmark & \cmark & \xmark \\
Field Latents~\cite{field} & \cmark & \cmark & \xmark & \xmark & \xmark \\
\midrule
HeatGen (Ours) & \cmark & \cmark & \cmark & \cmark & \cmark \\
\bottomrule
\end{tabular}}
\caption{Comparison with related methods. Surface neighbourhood: the denoiser aggregates along the surface rather than in 3D or image space. Mesh-independent resolution: the resolution of the state can be chosen independently of the mesh vertices. Heat-based forward: heat flow is the corruption process, not a network operator. HeatGen is the only method with all five properties.}
\label{tab:position}
\end{table}

\paragraph{Relation to heat kernel textures.} Heat Kernel Textures~\cite{hktex} represent appearance as a sum of anisotropic heat kernels fitted to a texture. The local unfolding of~\cite{localhk} evaluates such kernels without mesh eigensystems. We reuse this unfolding and its connection angles. A fitted kernel representation is not a convenient diffusion state, however: heat flow leaves its parametric family, noise on kernel parameters has no clear meaning in signal space, and the fit is not unique. Analysis coefficients avoid all three problems.

\paragraph{Concurrent work.} TaoTex~\cite{taotex} improves fine detail in native material generation through data construction and feature fusion, and Hitem3D~2.0~\cite{hitem3d} combines multi-view guidance with native texture generation. Both work in Euclidean latent spaces and are complementary to the surface state studied here.

\section{Data, Baselines and Metrics}
\label{supp:protocol}

\subsection{Benchmarks}
\label{supp:data}
Table~\ref{tab:datasets} summarizes the test sets. Re-extracted geometry is obtained by marching cubes at $512^3$, as in image-to-3D generators.

\begin{table}[tb]
\centering
\small
\setlength{\tabcolsep}{3.2pt}
\resizebox{\linewidth}{!}{\begin{tabular}{l ccc}
\toprule
& GSO-Remesh & I2M-Wild & ThinSet-2L \\
\midrule
Assets & 1,030 & 500 & 300 \\
Source & GSO scans & generated images & Objaverse \\
Geometry & MC re-extracted & generator output & MC re-extracted \\
Ground truth & yes & no & yes \\
Used for & LPIPS, KID & KID, study & Leak \\
\bottomrule
\end{tabular}}
\caption{Test sets. GSO-Remesh and ThinSet-2L have ground-truth textures and are re-extracted by marching cubes to resemble generator outputs, while I2M-Wild uses generated meshes as they are.}
\label{tab:datasets}
\end{table}

\paragraph{GSO-Remesh.} We compute an SDF of each Google Scanned Object, re-extract it by marching cubes and transfer the scan texture along normal rays. Fused regions, where a ray hits both sides, keep the colour of the nearer side. UVs are discarded. The input view is rendered from a random azimuth at $15^\circ$ elevation. The 16 held-out views lie on a sphere at elevations from $-30^\circ$ to $45^\circ$.

\paragraph{I2M-Wild.} The 500 images show single objects on clean backgrounds. They are generated with GPT-Image-2.5 from text prompts and balanced over garments, plants, footwear, furniture, toys, vehicles and household items. Half of the meshes come from the Hunyuan3D~2.1 shape model and half from the TRELLIS.2 shape stage, both run with default settings. Meshes are used as produced, with no repair. Every texturing method receives the same mesh, the same image and the camera returned by the shape generator.

\paragraph{ThinSet-2L.} We search held-out Objaverse assets for two-layer parts automatically. For every surface point we cast a ray along $-n$; if it hits another sheet within $2h$, the hit point is the point's opposite $x'$. Connected regions of such points form layer pairs, and we keep a pair if the mean colours of its two layers differ by more than 10 CIEDE2000.
After marching-cubes re-extraction, 34\% of pairs are separate components, 52\% are joined at a hem or rim and 14\% are fused. The mean layer separation is 0.6$h$. The input view is chosen to maximize the visible area of both layers.

\subsection{Baselines}
\label{supp:baselines}

\paragraph{Released systems.} Each is run at its published size with its default sampler and its own conditioning interface. TEXGen uses xatlas UVs at $1024^2$ and image and text conditioning, with the text from an image captioner. TRELLIS.2 runs its texture stage on our mesh, voxelized by its own encoder. Hunyuan3D-Paint~2.1 uses multi-view generation, baking and inpainting with default views at a 2K texture. NaTex uses 6144 latent tokens and default steps.

\paragraph{LaFiTe.} LaFiTe has no public checkpoint, so we retrain it from its paper on our training data, including the re-extracted half, with HeatGen's budget. Its autoencoder reaches 34.0\,dB reconstruction PSNR under the paper's protocol, against 34.62\,dB reported in~\cite{lafite}.

\paragraph{GOATex.} GOATex has no public code. Following its paper, we ray-cast hit levels from the default multi-view cameras, texture each visibility layer separately with the multi-view generator of Hunyuan3D-Paint~2.1, and blend the layers in UV space with visibility-confidence weights. GOATex therefore differs from Hunyuan3D-Paint only in its occlusion handling. On the paper's interior-texturing examples, our reimplementation textures 97\% of interior faces, against 61\% for Hunyuan3D-Paint. We tune the number of hit levels and the blending temperature with the same 20-trial budget as our own variants.

\paragraph{Controlled surface states.} These share HeatGen's data, budget, image features, visibility flags and global stream, and differ only in the state and its local operator. The \emph{point latent} places a NaTex-style colour latent of $16{,}384$ tokens at our anchors; local attention uses geodesic distances on the tufted cover, and each surface point decodes from latents reachable along the surface within $r_{\rm c}$. \emph{Vertex colours} live on a $\sim$200k-vertex proxy with discrete heat flow and DiffusionNet-style heat blocks~\cite{diffusionnet}, as in DoubleDiffusion~\cite{double}. We also give our LaFiTe retrain the same pixel-aligned conditioning, and train HeatGen with global image features only and on original meshes only.

\subsection{Metrics}
\label{supp:metricdefs}
\textbf{Masked LPIPS.} We compute LPIPS (VGG) with its spatial distance map, upsample the map to image resolution and average it over the mask. Masks are computed from geometry and the input camera, identically for all methods. Images where a mask covers fewer than 64 pixels are skipped for that mask. \textbf{Occluded mask:} pixels whose surface point is not visible from the input camera (z-buffer test). \textbf{Thin mask:} pixels whose surface point has another sheet within $2h$ along $\pm n$. Supp.~\ref{supp:metrics} sweeps the threshold. \textbf{KID:} computed with Inception features on all held-out views against ground-truth renders (GSO-Remesh), or against 8k renders of artist-made assets from a commercial store, none in Objaverse (I2M-Wild). Reported $\times10^3$. \textbf{Leak:} Eq.~\eqref{eq:leak}, computed on surface samples rather than images, so that occlusion does not bias it. \textbf{LSD:} log-spectral distance between radially averaged power spectra of generated and ground-truth renders. \textbf{Edge $\Delta E$ and overshoot:} CIEDE2000 error in a 4-pixel band around annotated decal and material boundaries, and the maximum excursion beyond the colour range of the two sides. \textbf{Statistics:} paired bootstrap with 10k resamples over assets, and over seeds where given.

\subsection{User Study}
\label{supp:study}
In the main study, 48 participants, 12 of them 3D artists, each judged 40 I2M-Wild assets, giving 1,920 votes. For each asset they saw the input image and turntable videos of the seven results in random order, with method names hidden, and picked the texture that best matches the input image and looks most plausible. In a second study, 24 further participants judged 30 assets each with turntables zoomed on a thin or contact region (720 votes). Table~\ref{tab:study} reports the share of votes for each method. The study protocol was reviewed by the institutional review board of New York University Abu Dhabi, and all participants gave informed consent before taking part.

\begin{table}[tb]
\centering
\small
\begin{tabular}{l cc}
\toprule
Method & Main study (\%) & Thin regions (\%) \\
\midrule
TEXGen & 4.1 & 6.8 \\
TRELLIS.2 & 7.3 & 4.2 \\
Hunyuan3D-Paint 2.1 & 17.8 & 12.5 \\
NaTex & 10.6 & 6.0 \\
LaFiTe & 9.4 & 8.1 \\
GOATex & 13.2 & 17.4 \\
\midrule
HeatGen (Ours) & \textbf{37.6} & \textbf{45.1} \\
\bottomrule
\end{tabular}
\caption{User study. Share of votes each method receives when participants pick the best of all seven results. HeatGen receives the largest share in both studies, and the gap widens when participants focus on thin and contact regions. Percentages are rounded independently.}
\label{tab:study}
\end{table}

\section{Derivations}
\label{supp:theory}

\subsection{Steerability}
Let a chart frame rotate by $\phi$, so that $\theta'=\theta-\phi$. Then $\psi'_{n,m}=g_\rho J_m(\kappa_n r)e^{\mathrm im(\theta-\phi)}=e^{-\mathrm im\phi}\psi_{n,m}$. Because $c=\langle\psi,F\rangle$ is conjugate-linear in $\psi$, $c'_{n,m}=e^{\mathrm im\phi}c_{n,m}$ (Eq.~\eqref{eq:gauge}). This is exact: the angular dependence is analytic, and the coefficient integral is taken over the surface, not over an angular grid. Truncating orders at $|m|\le M_n$ loses only the angular energy above $M_n$, 0.6\% on our data (Table~\ref{tab:checks}).

\subsection{Heat Closure}
\label{supp:closure}
By the Jacobi--Anger expansion,
\begin{equation*}
J_m(\kappa r)e^{\mathrm im\theta}=\frac{\mathrm i^{-m}}{2\pi}\int_0^{2\pi}e^{\mathrm im\alpha}e^{\mathrm i\kappa u_\alpha\cdot x}\,\mathrm d\alpha,
\end{equation*}
with $u_\alpha=(\cos\alpha,\sin\alpha)$. Each windowed plane wave $g_\rho(x)e^{\mathrm ik\cdot x}$ has a Fourier transform proportional to $e^{-\rho^2|\omega-k|^2/2}$. Heat flow on the plane multiplies it by $e^{-\tau|\omega|^2}$. Completing the square in $\omega$ gives a Gaussian of variance $1/\rho'^2$ centred at $k'=k\rho^2/\rho'^2$. Its amplitude is
\begin{equation*}
\tfrac{\rho^2}{\rho'^2}\exp\!\Big(-\tfrac{\tau\rho^2|k|^2}{\rho'^2}\Big),\qquad \rho'^2=\rho^2+2\tau,
\end{equation*}
which depends only on $|k|$. Transforming back and recombining over $\alpha$ gives Eq.~\eqref{eq:closure}. The order $m$ is unchanged, the window widens to $\rho'$, and the frequency drops to $\kappa'=\kappa\rho^2/\rho'^2$. The same computation in the optical setting gives the shape invariance of Bessel--Gauss beams~\cite{bessel}. Fig.~\ref{fig:atoms} checks the closed form against FFT heat flow on a $1024^2$ grid: the maximum relative error is below $10^{-10}$.

\subsection{Frame Bounds}
\label{supp:frame}
\paragraph{Discrete operators.} All operators act on fields sampled at the area-weighted quadrature points $q$ of Supp.~\ref{supp:impl}, with inner product $\langle F,G\rangle=\sum_q a_q\overline{F(q)}G(q)$. $S$ evaluates $\sum c_{i,n,m}\psi_{i,n,m}$ at the quadrature points, and $S^*$ is its adjoint in this inner product, which is Eq.~\eqref{eq:analysis} by quadrature. The singular values of $S$ have a sharp gap: about 63\% of them lie below $10^{-6}$ of the largest, and the rest lie in $[\sqrt A,\sqrt B]$. We take $V$ to be the span of the left singular vectors above the gap and treat the others as zero, so $A$ and $B$ are the extreme eigenvalues of $\Phi=SS^*$ on $V$. On this operator, $D=\Phi^{-1}S$ is the canonical dual decoder and $P=S^*\Phi^{-1}S$ the orthogonal projector onto the retained right singular vectors. Conjugate gradients started from zero compute $D$ at a rate set by $B/A$; the directions below the gap are not excited within the few iterations used.

On a regular anchor lattice, the frame operator $\Phi=SS^*$ commutes with lattice translations. Its spectrum is therefore the union over the Brillouin zone of the spectra of finite fibre matrices (Fourier fiberization). Each fibre matrix has size equal to the number of atoms per chart, and its entries are lattice sums of atom cross-spectra. We evaluate them numerically for rings $\kappa_n\rho=c_\Delta n$ and order cut-offs $M_n=\lceil r_m\kappa_n\rho\rceil$. Table~\ref{tab:frame} reports $B/A$ for several atom families. Wider windows lower $B/A$, but the number of quadrature points per chart, and hence the cost of analysis, grows roughly as $(\rho/h)^2$.

\begin{table}[tb]
\centering
\small
\begin{tabular}{ccc c}
\toprule
Window $\rho/h$ & Ring spacing $c_\Delta$ & Order cut-off $r_m$ & $B/A$ \\
\midrule
0.5 (ours) & 2.0 & 1.5 & 2.86 \\
0.6 & 2.0 & 1.5 & 2.76 \\
0.7 & 2.0 & 1.5 & 2.13 \\
0.5 & 2.5 & 2.0 & 4.56 \\
\bottomrule
\end{tabular}
\caption{Frame bounds on a regular anchor lattice for several atom families. Wider windows lower $B/A$, but the cost of analysis grows with the window area; denser rings raise it.}
\label{tab:frame}
\end{table}

On irregular anchors, $B/A$ is estimated by Lanczos iterations on $\Phi$ restricted to band-limited signals. Farthest-point sampling alone gives a median of 3.41, and five Lloyd iterations reduce it to 2.94. A low $B/A$ matters for two reasons: conjugate gradients converge quickly, and the token loss is a faithful proxy for the signal loss.

\subsection{Curvature}
In geodesic normal coordinates around $p_i$, the metric is $g=I+O(Kr^2)$ and the Laplace--Beltrami operator differs from the flat Laplacian by $O(K)$ terms. Over the support of a blurred atom, $r\lesssim3\rho'$, so the closed form of Eq.~\eqref{eq:closure} has relative error $O(K\rho'^2)$. At $\tau\le\tau_c$ this is $O(K\rho^2)$. With $\rho=h/2$ and anchors much denser than the curvature scale, the measured error is 0.6--2.7\% from low to high curvature (Table~\ref{tab:checks}).

\subsection{Adjoint Identity and Barriers}
Heat flow $H_\tau$ is self-adjoint in $L^2(\mathcal M,\mathrm dA)$, so $\langle\psi,H_\tau F\rangle=\langle H_\tau\psi,F\rangle$. This is Eq.~\eqref{eq:adjoint}. On a surface cut along $\mathcal B$ with Neumann conditions, the heat semigroup is again self-adjoint and preserves constants on every component. The identity therefore continues to hold, and the forward mean never moves colour across a barrier. For charts cut by $\mathcal B$, we compute $H_\tau\psi$ by implicit Euler on the chart patch and store it at 16 heat times.

\subsection{Anchor-Graph Heat}
The anchor-graph Laplacian $L=M_a^{-1}(D'-W')$, with $M_a=\mathrm{diag}(m)$, has the following properties. It is self-adjoint and positive semi-definite in the $M_a$ inner product; $L\mathbf 1=0$ and $m^\top L=0$; $m^\top u(\tau)$ is conserved under $u(\tau)=e^{-(\tau-\tau_c)L}u$. With $m_i=\int\bar w_i\,\mathrm dA$, $m^\top u$ equals the integral of the interpolated field $\sum_i\bar w_iu_i$, so the area-weighted mean colour is conserved exactly. If the graph restricted to a connected component is connected, $u(\tau)$ converges to that component's mean. The threshold of twice the covering radius guarantees this connectivity. The factor $1/(4\pi\delta^2)$ in $W'$ is the normalization of~\cite{belkin}: $(Lu)_i=(4\pi\delta^2)^{-1}\sum_j m_je^{-\bar d_{ij}^2/4\delta}(u_i-u_j)$ is their point-cloud estimate of $-\Delta$ with kernel width $\delta$, which scales as $s^{-2}$ when the surface is scaled by $s$. Hence $e^{-(\tau-\tau_c)L}$ advances the same heat time $\tau$ that the blurred atoms use below $\tau_c$. Graph heat equations have also been used directly as forward processes for graph signals~\cite{gad}. Here the graph only carries the coarse scales of a surface signal.

\subsection{Handoff}
\label{supp:handoff}
\paragraph{Choice of $\tau_c$.} At $\tau_c$ a blurred ring-1 atom has window $\rho'=\sqrt5\rho\approx1.1h$, well inside the chart radius $r_{\rm c}=3.4h$: less than 0.9\% of its energy lies beyond $r_{\rm c}$. The handoff is therefore not forced by the chart size. It is justified by frequency: every frequency that survives at $\tau_c$ satisfies $\kappa'h\le0.8<\pi$, below the anchor Nyquist rate, so anchor values can represent it. Continuing with blurred atoms would widen their support and their quadrature cost, which grows as $\rho'^2$.

\paragraph{Residual energy.} Anchor values are window-weighted local means, so they carry the order-0 part of the surviving signal and drop orders $m\ge1$. Consider a ring-1 atom ($\kappa_1\rho=2$). By Eq.~\eqref{eq:closure}, its order-$m\ge1$ components have already decayed strongly at $\tau_c=2\rho^2$, and an earlier handoff would drop more of them. On natural textures the dropped $m\ge1$ energy totals 1.6\% of the signal norm (Table~\ref{tab:tcsweep}). It is larger for textures that are strongly anisotropic at chart scale, up to 3.4\% on plaids. Even there our LSD stays within 0.04 of exact mesh heat, because the denoiser regenerates the dropped orders at finer scales.

\paragraph{Sensitivity to $\tau_c$.} Table~\ref{tab:tcsweep} varies the handoff time at M-scale and also reports the share of $m\ge1$ energy dropped at $\tau_c$. An early handoff ($\rho^2$) drops twice as much $m\ge1$ energy, most visibly on stripes and plaids, and costs KID. Later handoffs ($4\rho^2$, $8\rho^2$) drop less, but the operator becomes slower, and the wider blurred atoms reach across junctions more often, which raises Leak. $2\rho^2$ gives the best KID and Leak; $4\rho^2$ slightly improves LSD (1.51 versus 1.52), at higher cost and leakage.

\begin{table}[tb]
\centering
\small
\resizebox{\linewidth}{!}{\begin{tabular}{l ccccc}
\toprule
$\tau_c$ & KID$\downarrow$ & LSD$\downarrow$ & Leak$\downarrow$ & Dropped (\%) & Op (ms) \\
\midrule
$\rho^2$ & 15.4 & 1.56 & 0.043 & 3.1 & 7.2 \\
$2\rho^2$ (ours) & \textbf{15.1} & \underline{1.52} & \textbf{0.041} & 1.6 & 9.8 \\
$4\rho^2$ & 15.2 & \textbf{1.51} & 0.043 & 0.8 & 16.5 \\
$8\rho^2$ & 15.5 & 1.55 & 0.046 & 0.4 & 31.0 \\
\bottomrule
\end{tabular}}
\caption{Effect of the handoff time $\tau_c$ at M-scale. Dropped is the order-$m\ge1$ energy lost at the handoff. Earlier handoffs drop more energy, and later ones slow the operator and raise leakage; $2\rho^2$ is best on KID and Leak, while $4\rho^2$ has the lowest LSD.}
\label{tab:tcsweep}
\end{table}

\subsection{Ring Attenuation}
\label{supp:gamma}
The ring-wise loss weights use $\gamma_n(\tau)=\|H_\tau\psi_{n,m}\|/\|\psi_{n,m}\|$, averaged over $|m|\le3n$. By Eq.~\eqref{eq:closure} it has a closed form up to a one-dimensional radial integral. Ring 0 decays as $\rho/\rho'$. High rings decay much faster: by $\tau=\rho^2$, rings $n\ge3$ keep less than $10^{-5}$ of their energy. Because heat time is log-uniform, this allows the process to spend comparable effort on every scale. It is also why a diagonal schedule (ring heat decay in Table~\ref{tab:forwardfull}) can mimic only the decay of each ring, not the spatial coupling between rings.

\subsection{Sampling with Projected Noise}
\label{supp:noise}
The frame is redundant, so most coefficient vectors are not tokens of any texture: about 63\% of directions lie outside $\mathrm{range}(S^*)$. $P=S^*D=S^*\Phi^{-1}S$ is the orthogonal projector onto that range, and $P\epsilon$ is isotropic Gaussian noise on it, with covariance $P$.

Since $DP=D$, the decoded noise $DP\epsilon=\Phi^{-1}S\epsilon$ has covariance $\Phi^{-1}SS^*\Phi^{-1}=\Phi^{-1}$ on $V$, with spectrum in $[1/B,1/A]$.

The forward marginal is $\mathcal N(\alpha_t\mathcal A_{\tau_t}c_0,\sigma_t^2P)$, with a mean linear in $c_0$. DDIM's derivation needs only these two facts, with $\epsilon$ replaced by $P\epsilon$. The update in Alg.~\ref{alg:sample} is therefore the deterministic DDIM step for this process.

\subsection{Equivariance of the Denoiser}
Let every chart $i$ rotate its frame by $\phi_i$. Tokens transform as $z^{(m)}_i\mapsto e^{\mathrm im\phi_i}z^{(m)}_i$, and connection angles as $\phi_{ij}\mapsto\phi_{ij}+\phi_i-\phi_j$. Each layer is equivariant. \emph{Embedding and output maps} act separately on each order, by complex scalars that multiply rings and channels, so they commute with the phases. \emph{Attention values} are transported by $e^{\mathrm im\phi_{ij}}$. After the transformation, the transported value picks up exactly the phase $e^{\mathrm im\phi_i}$ of the receiving chart. \emph{Attention logits} use order-0 features and $\mathrm{Re}\langle q_i^{(m)},e^{\mathrm im\phi_{ij}}k_j^{(m)}\rangle$, which are invariant. \emph{Nonlinearities} scale each order by a function of invariants. \emph{The global stream} reads and writes only order-0 features. \emph{Image orientation} $e^{-\mathrm i\beta_i}$ gains $e^{\mathrm i\phi_i}$, as an order-1 token does.
Composition preserves equivariance. In single precision the measured error is $2\times10^{-6}$ (Table~\ref{tab:checks}). Errors in the connection angles themselves are not covered by this argument. They grow linearly with $m$, which is why only $|m|\le8$ is exchanged.

\subsection{Complexity}
Let $N$ be the number of anchors, $a=133$ the number of atoms per chart, $C$ the number of channels, and $n_q$ the number of quadrature points per chart (about $64\cdot\pi(r_{\rm c}/h)^2$). \emph{Analysis and synthesis} cost $O(Nn_qaC)$, and each is local to a chart. \emph{Decoding} adds a few conjugate-gradient iterations of $SS^*$. \emph{Forward mean, $\tau\le\tau_c$:} an analysis with blurred atoms, so the same cost. \emph{Forward mean, $\tau>\tau_c$:} a Chebyshev polynomial of degree $d$ on the anchor graph, costing $O(dN\bar k C)$, with $\bar k\approx36$ neighbours.
None of these terms depends on the number of mesh vertices, which enters only once, through chart construction. Mesh-resolution heat instead costs $O(d|V|C)$ per evaluation, with $|V|\gg N$. Because the predicted texture changes, this cost recurs at every sampling step.

\section{Numerical Checks}
\label{supp:checks}

Table~\ref{tab:checks} collects the numerical checks referred to above. Forward-mean errors are relative $L^2$ errors on $\sim$1M-vertex spheres, tori and scans against implicit-Euler cotangent heat flow, frame bounds are measured over GSO-Remesh meshes, and transport errors are measured against the vector heat method~\cite{vectorheat}.

\begin{table}[tb]
\centering
\small
\resizebox{\linewidth}{!}{\begin{tabular}{l c}
\toprule
Check & Error \\
\midrule
\multicolumn{2}{l}{\emph{Approximations}} \\
Heat closure vs.\ FFT heat flow (max.\ relative) & $<$$10^{-10}$ \\
Forward mean vs.\ mesh heat, low curvature (\%) & 0.6 \\
Forward mean vs.\ mesh heat, high curvature (\%) & 2.7 \\
Angular truncation (\% of energy) & 0.6 \\
Neumann responses, interpolation (\%) & 0.3 \\
\midrule
\multicolumn{2}{l}{\emph{Frame}} \\
$B/A$, regular grid & 2.86 \\
$B/A$, our anchors (median / 95th percentile) & 2.94 / 3.21 \\
\midrule
\multicolumn{2}{l}{\emph{Equivariance}} \\
Gauge law, phase error & $4\times10^{-16}$ \\
Network under per-chart frame changes (float32) & $2\times10^{-6}$ \\
Transport phase error, $m=1$ / $8$ ($^\circ$, median) & 1.9 / 14.8 \\
\bottomrule
\end{tabular}}
\caption{Numerical checks of the approximations used in HeatGen. References are FFT heat flow on a fine grid, implicit-Euler mesh heat and the vector heat method.}
\label{tab:checks}
\end{table}

\section{Evaluation Details}
\label{supp:metrics}

\subsection{Thin-Mask Threshold}
\label{supp:sweep}
Table~\ref{tab:sweep} varies the sheet-distance threshold that defines LPIPS-thin from $0.5h$ to $8h$; $2h$ is about 0.8\% of the bounding-box diagonal. Over this range, the mask grows from 0.6\% to 13.5\% of held-out pixels. HeatGen has the lowest LPIPS-thin at every threshold, and the gap shrinks as the mask includes regions that are not thin.

\begin{table}[tb]
\centering
\small
\resizebox{\linewidth}{!}{\begin{tabular}{l ccccc}
\toprule
Threshold & $0.5h$ & $h$ & $2h$ & $4h$ & $8h$ \\
\midrule
\multicolumn{6}{l}{\emph{Mask coverage}} \\
Pixels (\%) & 0.6 & 1.4 & 3.1 & 6.8 & 13.5 \\
Assets & 142 & 247 & 371 & 566 & 802 \\
\midrule
\multicolumn{6}{l}{\emph{LPIPS-thin}$\downarrow$} \\
Hunyuan3D-Paint 2.1 & 0.236 & 0.231 & 0.224 & 0.216 & 0.207 \\
NaTex & 0.243 & 0.236 & 0.227 & 0.218 & 0.208 \\
LaFiTe & 0.231 & 0.226 & 0.219 & 0.212 & 0.205 \\
GOATex & 0.227 & 0.222 & 0.217 & 0.210 & 0.203 \\
HeatGen (Ours) & \textbf{0.201} & \textbf{0.199} & \textbf{0.198} & \textbf{0.197} & \textbf{0.194} \\
\bottomrule
\end{tabular}}
\caption{Sensitivity to the thin-mask threshold. LPIPS-thin on GSO-Remesh as the sheet-distance threshold varies; the $2h$ column corresponds to Table~\ref{tab:i2m}. HeatGen is best at every threshold.}
\label{tab:sweep}
\end{table}

\subsection{Other Metrics on the Same Masks}
\label{supp:masked}
Table~\ref{tab:masked} adds CLIP-I similarity to the input image, PSNR and KID on occluded regions, and diversity, measured as the mean LPIPS on occluded regions between four samples with different seeds. HeatGen has the best occluded KID, while its diversity (0.124) is second to NaTex (0.126). Hunyuan3D-Paint has the best CLIP-I, reflecting its fidelity in the input view.

\begin{table}[tb]
\centering
\small
\setlength{\tabcolsep}{3.2pt}
\resizebox{\linewidth}{!}{\begin{tabular}{l cccc}
\toprule
Method & CLIP-I$\uparrow$ & PSNR-occ$\uparrow$ & KID-occ$\downarrow$ & Div.-occ$\uparrow$ \\
\midrule
Hunyuan3D-Paint 2.1 & \textbf{0.902} & 19.8 & 14.6 & 0.118 \\
NaTex & 0.893 & 20.4 & 13.8 & \textbf{0.126} \\
LaFiTe & 0.891 & 20.3 & 13.9 & 0.121 \\
GOATex & 0.897 & 20.6 & 13.2 & 0.122 \\
HeatGen (Ours) & \underline{0.899} & \textbf{21.2} & \textbf{12.4} & \underline{0.124} \\
\bottomrule
\end{tabular}}
\caption{Additional metrics on GSO-Remesh. HeatGen has the best occluded KID; NaTex has the highest diversity and HeatGen is second. Hunyuan3D-Paint has the best CLIP-I.}
\label{tab:masked}
\end{table}

\subsection{Leakage}
\label{supp:leak}
After marching-cubes re-extraction, ThinSet-2L layer pairs are separate components (34\%), joined at a hem or rim (52\%) or fused into one sheet (14\%). Fused pairs are excluded, because no surface colour field can give one sheet two colours. Table~\ref{tab:leak} splits Leak by topology and by geodesic distance to the nearest junction. HeatGen leaks mainly near junctions, where charts necessarily span both layers, yet it stays below every baseline there; away from junctions it is less than half that of the best baseline. The cascade's smaller charts reduce Leak within $2h$ of a junction to 0.052.

\begin{table}[tb]
\centering
\small
\resizebox{\linewidth}{!}{\begin{tabular}{l c cc cc}
\toprule
& & \multicolumn{2}{c}{Topology} & \multicolumn{2}{c}{To junction} \\
\cmidrule(lr){3-4}\cmidrule(lr){5-6}
Method & All & separate & joined & ${<}2h$ & ${>}4h$ \\
\midrule
TEXGen & 0.09 & 0.08 & 0.10 & 0.11 & 0.08 \\
Hunyuan3D-Paint 2.1 & 0.12 & 0.10 & 0.13 & 0.15 & 0.11 \\
NaTex & 0.25 & 0.23 & 0.26 & 0.28 & 0.24 \\
LaFiTe & 0.13 & 0.12 & 0.14 & 0.16 & 0.12 \\
GOATex & 0.07 & 0.06 & 0.08 & 0.09 & 0.06 \\
\midrule
HeatGen (Ours) & \textbf{0.036} & \textbf{0.012} & \textbf{0.051} & \textbf{0.062} & \textbf{0.029} \\
\bottomrule
\end{tabular}}
\caption{Leak on ThinSet-2L by layer topology and by geodesic distance to the nearest junction. All methods leak more where layers are joined, but HeatGen stays lowest everywhere and leaks little away from junctions.}
\label{tab:leak}
\end{table}

\paragraph{Reference values.} Leak measures the direction of the error rather than its size. Ground truth has Leak 0.00 and layer-swapped ground truth 1.00. Smoothing does not lower it: a constant mean colour per part gives 0.48 and heavy blur 0.33, because averaging two layers moves each towards the other.

\paragraph{CLOTH4D.} We also evaluate on 200 outfits from the test split of CLOTH4D~\cite{cloth4d}, whose garment layers are designed and textured separately. We use them as authored, without re-extraction, and keep outfits with two overlapping garments. Table~\ref{tab:cloth4d} shows the same ordering as on ThinSet-2L; the geodesic point latent is the controlled surface state of Supp.~\ref{supp:unique}.

\begin{table}[tb]
\centering
\small
\setlength{\tabcolsep}{3.2pt}
\begin{tabular}{l ccc}
\toprule
CLOTH4D & LPIPS-occ$\downarrow$ & LPIPS-thin$\downarrow$ & Leak$\downarrow$ \\
\midrule
Hunyuan3D-Paint 2.1 & 0.201 & 0.233 & 0.12 \\
NaTex & 0.197 & 0.238 & 0.27 \\
GOATex & 0.193 & 0.224 & 0.06 \\
Geodesic point latent & 0.191 & 0.219 & 0.049 \\
HeatGen (Ours) & \textbf{0.178} & \textbf{0.203} & \textbf{0.021} \\
\bottomrule
\end{tabular}
\caption{Results on CLOTH4D. Garment layers are used as authored, without re-extraction. The ordering of methods matches that on ThinSet-2L.}
\label{tab:cloth4d}
\end{table}

\section{Controlled and Scaling Studies}
\label{supp:studies}

\subsection{Forward Process}
\label{supp:replication}
Table~\ref{tab:forwardfull} lists the values behind Fig.~\ref{fig:forward} together with four further variants. Ring heat decay uses the closed-form attenuation of each ring without spatial coupling. Atoms + eigenbasis combines our atoms at fine scales with 1,024 Laplacian eigenvectors at coarse scales. The last variant runs mesh heat on a 50k-face decimated proxy. Caching the mesh-heat training targets removes the operator cost from training but not from sampling, and needs 10\,TB of storage. Steps and Time are defined as in Fig.~\ref{fig:forward}.

At full scale, the mean KID over three independent training seeds is 10.5, compared with 10.9 for per-ring + graph heat and 10.7 for mesh heat, while sampling in 29.1\,s instead of 86.3\,s.

\begin{table}[tb]
\centering
\small
\resizebox{\linewidth}{!}{\begin{tabular}{l cccc}
\toprule
Forward process & KID$\downarrow$ & LSD$\downarrow$ & Iters (\%)$\downarrow$ & Time (s)$\downarrow$ \\
\midrule
Gaussian (VP) & 16.8 & 1.86 & 100 & 11.2 \\
Per-ring schedule & 16.0 & 1.69 & 74 & 9.0 \\
Ring heat decay & 16.3 & 1.73 & 81 & 10.1 \\
Per-ring + graph heat & 15.4 & 1.57 & 61 & 6.8 \\
Atoms + eigenbasis & \underline{15.2} & \underline{1.53} & \underline{57} & \underline{5.4} \\
Mesh heat & 15.3 & 1.55 & 58 & 36.2 \\
Mesh heat, decimated & 15.6 & 1.61 & 63 & 10.2 \\
\midrule
Ours & \textbf{15.1} & \textbf{1.52} & \textbf{55} & \textbf{4.9} \\
\bottomrule
\end{tabular}}
\caption{All forward-process variants at M-scale. Besides the quantities in Fig.~\ref{fig:forward}, we report the log-spectral distance (LSD).}
\label{tab:forwardfull}
\end{table}

\subsection{Complete Ablations}
\label{supp:ablation}
Table~\ref{tab:fullabl} lists every ablation at M-scale. The variant without gauge equivariance uses one smooth tangent field per mesh, computed with the vector heat method~\cite{vectorheat}, and a non-equivariant network. Such a field has at least two singularities on a genus-0 surface, and charts near them receive unstable frames. Fitted coefficients (least squares with Cholesky whitening) are slightly worse and slower than analysis coefficients. Without noise projection, the network spends capacity on the roughly 63\% of directions that are not valid tokens.

On the network side, exchanging only $|m|\le2$ between charts is clearly worse than $|m|\le8$. Euclidean attention reintroduces leakage; a normal-based gate removes about half of it but does not close the gap. Removing the global stream or its 3D positions lowers global coherence and slightly raises leakage, and 64 global tokens are too few. Smaller charts are the only variant with lower Leak than the full model, at a clear cost in KID; the cascade exploits the same effect at fine scales only (Supp.~\ref{supp:more}). Ring-wise min-SNR weighting and Lloyd relaxation each improve KID by a few tenths; without Lloyd relaxation, $B/A$ rises from 2.94 to 3.41.

\begin{table}[tb]
\centering
\small
\resizebox{\linewidth}{!}{\begin{tabular}{l cccc}
\toprule
Variant & KID$\downarrow$ & LPIPS-occ$\downarrow$ & LPIPS-thin$\downarrow$ & Leak$\downarrow$ \\
\midrule
Full model & 15.1 & 0.191 & 0.214 & 0.041 \\
\midrule
\multicolumn{5}{l}{\emph{State}} \\
Fourier--Bessel $\to$ Gabor atoms & 16.0 & 0.197 & 0.221 & 0.046 \\
Analysis $\to$ fitted coefficients & 15.6 & 0.195 & 0.219 & 0.043 \\
w/o noise projection & 15.8 & 0.196 & 0.220 & 0.042 \\
Chart tokens $\to$ vertex colours & 16.6 & 0.199 & 0.224 & 0.049 \\
Chart tokens $\to$ point latent (Euclidean) & 15.4 & 0.199 & 0.233 & 0.187 \\
\midrule
\multicolumn{5}{l}{\emph{Network}} \\
w/o gauge equivariance & 16.3 & 0.198 & 0.223 & 0.044 \\
Transport $|m|\le8\to|m|\le2$ & 15.6 & 0.195 & 0.218 & 0.041 \\
Geodesic $\to$ Euclidean attention & 15.2 & 0.192 & 0.229 & 0.168 \\
Chart radius $3.4h\to2.4h$ & 15.5 & 0.194 & 0.213 & 0.036 \\
Chart radius $3.4h\to4.4h$ & 15.2 & 0.192 & 0.217 & 0.047 \\
w/o global stream & 16.7 & 0.201 & 0.218 & 0.044 \\
Global stream w/o 3D positions & 15.5 & 0.193 & 0.215 & 0.043 \\
Global tokens $256\to64$ & 15.6 & 0.194 & 0.215 & 0.042 \\
\bottomrule
\end{tabular}}
\caption{Full ablation study at M-scale. Rows are grouped by the part of HeatGen they change, and all variants use the same data, budget and image conditioning.}
\label{tab:fullabl}
\end{table}

\subsection{Surface States at Full Scale}
\label{supp:unique}
Table~\ref{tab:states} compares surface states trained with HeatGen's data, budget and pixel-aligned conditioning (Supp.~\ref{supp:baselines}). Geodesic attention and decoding already help a point latent, and enlarging its latent fourfold or giving each token a learned tangent frame changes little, because without exact steerability the frames are not consistent across charts. Chart tokens improve on every variant. Chart tokens are also less sensitive to the triangulation: across five triangulations of the same shape, generations vary by 0.031 LPIPS, against 0.047 for the point latent and 0.094 for vertex colours, because the analysis has a fixed decoder. Giving LaFiTe our conditioning helps it, but it still leaks, and HeatGen with global image features only remains ahead of every baseline in Table~\ref{tab:i2m} on occluded and thin regions and on Leak.

\begin{table}[tb]
\centering
\small
\resizebox{\linewidth}{!}{\begin{tabular}{l cccc}
\toprule
State & LPIPS-occ$\downarrow$ & LPIPS-thin$\downarrow$ & KID$\times10^3$$\downarrow$ & Leak$\downarrow$ \\
\midrule
\multicolumn{5}{l}{\emph{Other surface states}} \\
Point latent, geodesic & 0.186 & 0.214 & 11.9 & 0.061 \\
\quad $4\times$ latent tokens & 0.184 & 0.212 & 11.8 & 0.058 \\
\quad + learned tangent frames & 0.182 & 0.210 & 11.7 & 0.057 \\
Vertex colours & 0.190 & 0.218 & 12.8 & 0.046 \\
LaFiTe + our conditioning & 0.187 & 0.216 & 11.2 & 0.13 \\
\midrule
\multicolumn{5}{l}{\emph{HeatGen variants}} \\
w/o pixel-aligned features & 0.175 & 0.201 & 11.6 & 0.037 \\
w/o re-extracted training meshes & 0.174 & 0.202 & 11.9 & 0.038 \\
\midrule
HeatGen (Ours) & \textbf{0.171} & \textbf{0.198} & \textbf{10.5} & \textbf{0.036} \\
\bottomrule
\end{tabular}}
\caption{Comparison of surface states at full scale. All states are trained with HeatGen's data, budget and pixel-aligned conditioning; HeatGen variants remove one part of the data or conditioning. HeatGen is best on every metric.}
\label{tab:states}
\end{table}

\subsection{Sensitivity}
\paragraph{Number of anchors.} KID is 12.9 at $N=4{,}096$, reaches its lowest value of 10.5 at our choice of 16,384 anchors, and rises to 11.2 at 32,768 anchors. Time per asset grows roughly linearly, from 14 to 66\,s, because local attention has a fixed neighbourhood.

\paragraph{Sampling steps.} With 25 instead of 50 DDIM steps, KID rises from 10.5 to 11.8 and time per asset falls from 37.0 to 22.5\,s.

\paragraph{Redundancy.} Keeping only every other order shrinks the state but raises $B/A$ to 7.9, which slows decoding and raises KID from 15.1 to 15.5 at M-scale.

\begin{figure*}[!t]
\centering
\begin{minipage}[t]{0.615\textwidth}\vspace{0pt}\centering
\includegraphics[width=\linewidth]{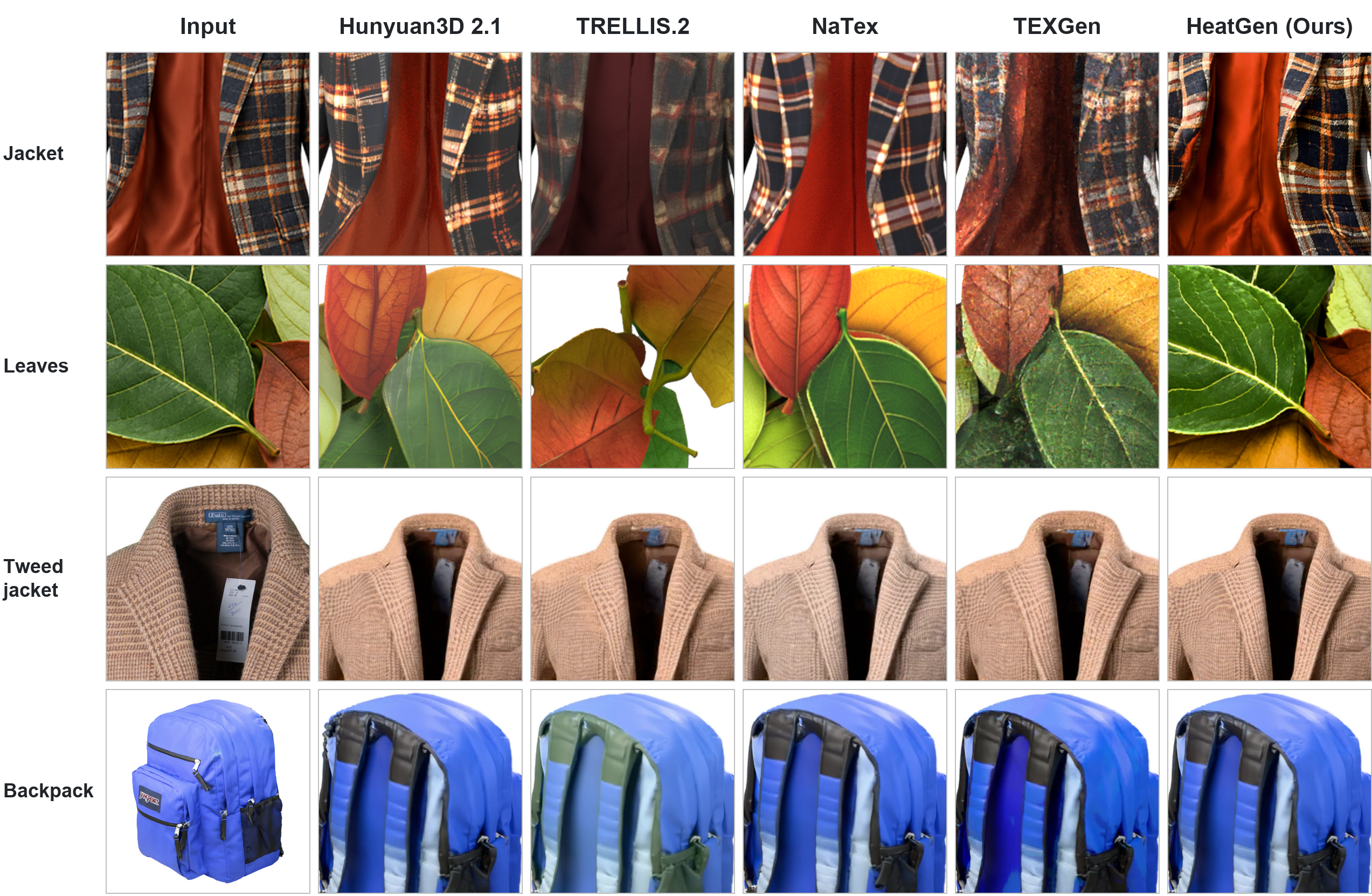}
\subcaption{Thin and contact regions}\label{fig:thinclose}
\end{minipage}\hfill
\begin{minipage}[t]{0.37\textwidth}\vspace{0pt}\centering
\includegraphics[width=\linewidth]{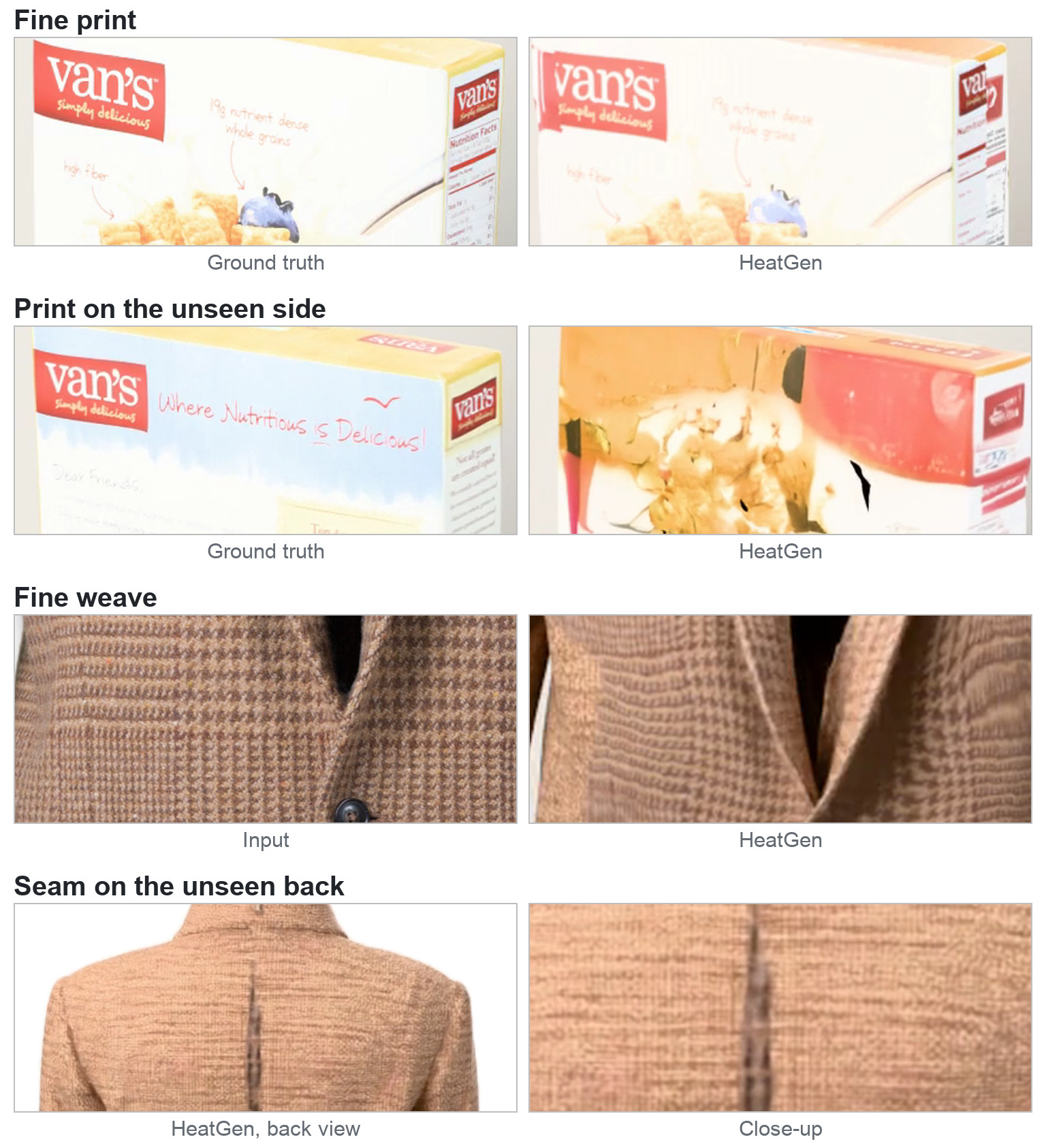}
\subcaption{Failure cases}\label{fig:failures}
\end{minipage}
\caption{Additional qualitative results. (a) Close-ups of thin and contact regions for the methods of Fig.~\ref{fig:qual}. (b) Typical failure cases of HeatGen: fine print is blurred, print on the unseen side is hallucinated, a fine regular weave is rendered coarser, and a seam can appear on the unseen back.}
\label{fig:qualsupp}
\end{figure*}

\section{Additional Results}
\label{supp:more}

\paragraph{Repainting scans.} We mask regions of 240 scanned GSO and OmniObject3D~\cite{omni} objects and repaint them conditioned on the remaining texture. A barrier on the mask boundary keeps known colours from bleeding inward through the forward process and generated colours from bleeding outward through attention. It raises PSNR in a $2h$ band around the boundary by 7.3\,dB.

\paragraph{Transfer across discretizations.} We generate a texture once and transfer it to a 10\% QEM simplification, an isotropic remesh and a point cloud by re-decoding at the new points. Renders before and after transfer agree to 38.9, 40.2 and 37.6\,dB PSNR, while resampled vertex colours reach 29.4, 31.8 and 27.2\,dB.

\paragraph{Cascade.} The local-only refinement model uses $65{,}536$ anchors (Sec.~\ref{sec:training}), doubling texture resolution from about $900^2$ to $1800^2$ texels and halving chart width. LPIPS and layer separation improve, with KID 10.3\ (Table~\ref{tab:cascade}); gains are larger near junctions, where smaller charts cross layers less often, and on printed text. On 200 text and logo assets, OCR character accuracy rises from 48\% to 69\%, close to Hunyuan3D-Paint's 71\% with a 2K texture. The cascade also halves slight ringing at sharp colour edges.

\begin{table}[tb]
\centering
\small
\resizebox{\linewidth}{!}{\begin{tabular}{l cccccc}
\toprule
& LPIPS$\downarrow$ & LPIPS-thin$\downarrow$ & KID$\times10^3$$\downarrow$ & Leak$\downarrow$ & Leak${<}2h$$\downarrow$ & Time (s)$\downarrow$ \\
\midrule
Single level & 0.135 & 0.198 & 10.5 & 0.036 & 0.062 & \textbf{37} \\
+ cascade & \textbf{0.132} & \textbf{0.193} & \textbf{10.3} & \textbf{0.034} & \textbf{0.052} & 78 \\
\bottomrule
\end{tabular}}
\caption{Effect of the cascade. LPIPS and KID on GSO-Remesh, Leak on ThinSet-2L overall and within $2h$ of a layer junction. The second level improves LPIPS and layer separation, most clearly near junctions, at about twice the sampling time.}
\label{tab:cascade}
\end{table}

\paragraph{Runtime.} For one asset at $N=16{,}384$ with 50 steps on one H100, chart construction takes about 4.6\,s, sampling 29.1\,s and the final decode 3.3\,s, 37.0\,s in total. Most of the sampling time is spent in the network; the conjugate-gradient decode and the forward mean take about a quarter of each step.

\paragraph{Chart construction on difficult inputs.} Construction succeeds when every anchor's chart covers at least 90\% of its disc. This holds for 100\% of raw generator outputs and 99.4\% of raw Objaverse triangle soups; the few failures come from degenerate, zero-area fans.

\paragraph{Operating limits.} Separate surfaces never share charts, even when close or self-intersecting, because unfolding follows connectivity. Layers joined at a hem or rim share charts within about $r_{\rm c}$ of the junction (Table~\ref{tab:leak}). Slivers distort hinge rotations; their effect is bounded by the transport errors in Table~\ref{tab:checks}, evaluated on generated meshes containing slivers.

\section{Qualitative Results}
\label{supp:qual}

Fig.~\ref{fig:thinclose} examines thin and contact regions against the baselines of Fig.~\ref{fig:qual}. Fig.~\ref{fig:failures} shows where the method fails. Fine print and fine weaves are limited by the anchor spacing (Sec.~\ref{sec:limitations}), and the unseen side is not determined by the input image.

\section{Broader Impact}
\label{supp:impact}
HeatGen can lower the cost of texturing generated and scanned 3D assets for games, film and AR. It can also facilitate imitation of copyrighted textures or brand marks. Training uses assets licensed for research. ThinSet-2L is released as asset identifiers and processing scripts, respecting source licences. Full training takes 160 H100-days, comparable to recent native texture generators; sampling runs on one GPU.

\end{document}